\documentclass[10pt,twocolumn,letterpaper]{article}

\usepackage[algorithms]{wacv}      
\usepackage{graphicx}
\usepackage{amsmath}
\usepackage{amssymb}
\usepackage{booktabs}

\definecolor{wacvblue}{rgb}{0.21,0.49,0.74}
\usepackage[pagebackref,breaklinks,colorlinks,allcolors=wacvblue]{hyperref}

\def\wacvPaperID{437} 
\def\confName{WACV}
\def\confYear{2026}

\usepackage[dvipsnames]{xcolor}

\definecolor{myred}{RGB}{220, 20, 60}

\title{OMeGa: Joint Optimization of Explicit Meshes and Gaussian Splats for Robust Scene-Level Surface Reconstruction}

\author{
Yuhang Cao$^{1}$\thanks{Equal contribution.} \quad
Haojun Yan$^{2}$\footnotemark[1] \quad
Danya Yao$^{1}$\thanks{Corresponding author.} \\
$^1$ Tsinghua University, China \quad
$^2$ Beihang University, China\\
{\tt\small caoyh24@mails.tsinghua.edu.cn \quad haojunyan@buaa.edu.cn \quad yaody@tsinghua.edu.cn}
}

\begin{document}
\maketitle
\begin{abstract}
Neural rendering with Gaussian splatting has advanced novel view synthesis, and most methods reconstruct surfaces via post-hoc mesh extraction. However, existing methods suffer from two limitations: (i) inaccurate geometry in texture-less indoor regions, and (ii) the decoupling of mesh extraction from optimization, thereby missing the opportunity to leverage mesh geometry to guide splat optimization. In this paper, we present OMeGa, an end-to-end framework that jointly optimizes an explicit triangle mesh and 2D Gaussian splats via a flexible binding strategy, where spatial attributes of Gaussian Splats are expressed in the mesh frame and texture attributes are retained on splats.
To further improve reconstruction accuracy, we integrate mesh constraints and monocular normal supervision into the optimization, thereby regularizing geometry learning.
In addition, we propose a heuristic, iterative mesh-refinement strategy that splits high-error faces and prunes unreliable ones to further improve the detail and accuracy of the reconstructed mesh.
OMeGa achieves state-of-the-art performance on challenging indoor reconstruction benchmarks, reducing Chamfer-$L_1$ by 47.3\% over the 2DGS baseline while maintaining competitive novel-view rendering quality. The experimental results demonstrate that OMeGa effectively addresses prior limitations in indoor texture-less reconstruction.

\end{abstract}
    
\section{Introduction}
\label{sec:intro}
Novel View Synthesis and 3D Reconstruction are two fundamental tasks in computer vision and graphics, with widespread applications in virtual/augmented reality, autonomous driving, and robotics. Recent advancements in neural rendering have provided promising solutions for novel view synthesis and 3D Reconstruction. Neural Radiance Fields (NeRF)\cite{mildenhall2021nerf} utilizes a multi-layer perceptron (MLP) to learn volume density and emitted radiance from spatial location and viewing direction of camera pose implicitly. Following works\cite{wang2021neus,fu2022geo,wang2022hf,yariv2023bakedsdf} integrate signed distance functions (SDFs) into the NeRF framework to represent accurate surface for scene reconstruction. 3D Gaussian Splatting (3DGS)\cite{kerbl20233d}, which utilizes explicit 3D ellipsoids as scene representation, has shown great potential due to its fast training time and real-time realistic novel views rendering. To further improve the reconstruction quality, following works\cite{guedon2024sugar, huang20242d} have proposed to flatten the 3D ellipsoids and transform 3D splats into 2D surfels.

\begin{figure}[htb]
  \centering
  \begin{subfigure}[b]{0.46\linewidth}
    \centering
    \includegraphics[width=\linewidth]{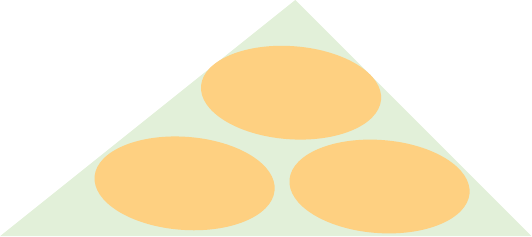}
    \caption{Previous Works}
    \label{fig:other binding}
  \end{subfigure}
  \hfill
  \begin{subfigure}[b]{0.46\linewidth}
    \centering
    \includegraphics[width=\linewidth]{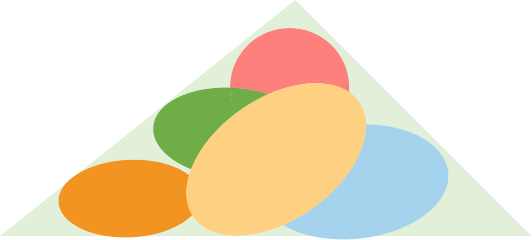}
    \caption{OMeGa}
    \label{fig:ours binding}
  \end{subfigure}
  \caption{\textbf{Different Binding Strategies.} (a) Previous works allocate a fixed number of splats with predefined positions, scales, and orientations. (b) OMeGa adopts a flexible binding strategy to enhance splats’ fitting capability.}
  \label{fig:binding}
\end{figure}

While significant progress has been made in object-level surface reconstruction, indoor scene reconstruction remains particularly challenging due to complex object arrangements and large texture-less regions. These factors often lead to degraded reconstruction quality in both geometry and appearance. Additionally, most Gaussian Splatting methods perform mesh extraction independently after training. This pipeline overlooks the potential of leveraging mesh geometry as a constraint during optimization, which could guide the learning of Gaussian Splats in turn. In this work, we propose a novel, end-to-end differentiable hybrid representation that integrates splats with meshes for indoor scene reconstruction, which simultaneously achieves high-fidelity rendered images and accurate meshes.

Previous works\cite{qian2024gaussianavatars,waczynska2024games,choi2024meshgs} have explored hybrid representations of Gaussian Splats and geometry representations such as SDFs and meshes. However, as shown in \cref{fig:other binding}, attributes of splats in these methods are predefined based on the mesh faces, thereby severely constraining the expressive power of Gaussian Splatting. In contrast, our flexible binding strategy, which allows splats to move freely within the plane and to be duplicated or removed, can guarantee strong expressiveness. Furthermore, most existing methods\cite{guedon2024sugar, lin2025directlearningmeshappearance} rely on a detailed mesh extracted from a pretrained Gaussian Splatting model for initialization, typically involving a two-stage training pipeline. Leveraging a carefully designed mesh refinement strategy, our method can start from a quite coarse and noisy mesh, substantially improving efficiency without sacrificing final reconstruction quality.

 Specifically, we introduce a framework where 2D Gaussian primitives are flexibly bound to mesh faces, enabling a joint optimization of both representations. Then, by incorporating mesh representation directly into the optimization, we propose a mesh refinement strategy, including an error-based mesh subdivision method and a GS-based mesh removal method, which significantly improve the details and accuracy of the reconstructed mesh. Additionally, we impose mesh-based constraints on the mesh representation, which in turn helps regulate the 2D splats. The results demonstrate that our method achieves state-of-the-art performance on challenging benchmarks, effectively addressing limitations of previous approaches, such as geometry degradation in texture-less areas.

In summary, our contributions are as follows:
\begin{itemize}
    \item We propose a novel end-to-end joint-representation framework, which flexibly binds 2D splats to mesh and can differentiability optimize geometry and rendering simultaneously from scratch.
    \item By introducing an explicit mesh representation, we enable an iterative mesh refinement strategy and mesh-guided optimization of Gaussian Splats through geometry-aware constraints, significantly enhancing geometric accuracy and rendering quality.
    \item Extensive experiments demonstrate that our approach achieves state-of-the-art performance on rendering and reconstruction tasks, while demonstrating robustness to significant view changes, texture-less regions, and lighting variations.
\end{itemize}
\section{Related Work}
\label{sec:related}
\subsection{Neural Radiance Field}
Neural Rendering has experienced advancement since the Neural Radiance Fields (NeRF)\cite{mildenhall2021nerf}, which employs a multi-layer perceptron (MLP) to model the radiance field and utilizes volumetric rendering to produce high-fidelity rendered images. Following works focus on further improving rendering quality through refined ray sampling technique\cite{barron2021mip,barron2023zip}, and accelerating training and rendering speed through multi-resolution hash encoding\cite{muller2022instant}, baking technique\cite{yu2021plenoctrees, garbin2021fastnerf}, and less MLP\cite{fridovich2022plenoxels,sun2022direct}. However, the NeRF model learns a volume density field,  from which extracting a high-quality surface representation is challenging. 

Research on NeRF's geometry reconstruction has explored utilizing occupancy grid\cite{oechsle2021unisurf} and signed distance functions (SDFs)\cite{wang2021neus,fu2022geo,wang2022hf,yariv2023bakedsdf} to enhance the surface reconstruction quality. Following research\cite{rakotosaona2024nerfmeshing, tang2022nerf2mesh} have designed end-to-end pipelines that produce mesh and appearance model simultaneously. However, these mesh reconstruction methods tend to produce over-smooth results and have not been thoroughly explored in terms of effectiveness at the scene level.

\subsection{Gaussian Splatting}
Recently, 3DGS\cite{kerbl20233d} has attracted wide research interests due to its high-quality rendering images and real-time rendering speed. 3DGS explicitly utilizes 3D ellipsoids (Gaussian Splats) as the scene representation and employs the alpha-blending technique for rasterization. However, 3DGS exhibits suboptimal performance in scene reconstruction due to its unstructured representations. GOF\cite{yu2024gaussian} forms a Gaussian opacity field from 3D Gaussians for geometry extraction. Following methods, such as SuGaR\cite{guedon2024sugar} and PGSR\cite{chen2024pgsr}, utilize regularization and flatten 3D Gaussians into 2D Gaussians for better reconstruction results. 2DGS\cite{huang20242d} proposes to replace 3D ellipsoids with 2D surfels to achieve better surface reconstruction.
 
Vanilla Gaussian Splatting models encounter significant challenges in indoor scene reconstruction tasks, since they are more complex than object-level reconstruction and contain a large number of lox-texture regions and intricate details. Most research\cite{turkulainen2024dn,ren2024ags,wu2024surface,zhang20242dgs,xiang2024gaussianroom} utilizes monocular or sensor geometry cues as supervision to improve Gaussian Splatting's reconstruction quality for indoor room scenes.

Recent works\cite{qian2024gaussianavatars,waczynska2024games,choi2024meshgs} have focused on integrating geometry representations, such as SDFs and meshes, into the Gaussian Splatting model. GaussianAvatars\cite{qian2024gaussianavatars} proposes a method for avatar modeling that combines 3DGS with a parameterized deformable face model. GaMeS\cite{waczynska2024games} presents a rendering method that tightly couples splats with meshes by parameterizing each Gaussian primitive as a function of the mesh face vertices, allowing splats to adapt automatically with mesh deformations. However, attributes of splats in these methods are predefined based on the mesh faces, thereby severely constraining the expressive power of Gaussian Splatting. Furthermore, most existing methods\cite{guedon2024sugar, lin2025directlearningmeshappearance} rely on a detailed mesh extracted from a pretrained Gaussian Splatting model for initialization, typically involving a two-stage training pipeline. Unlike prior methods, our method maintains high-quality reconstruction even when initialized with coarse and noisy meshes, thanks to a flexible binding mechanism and an effective mesh refinement strategy. 

\section{Methods}

\begin{figure*}
  \includegraphics[width=\linewidth]{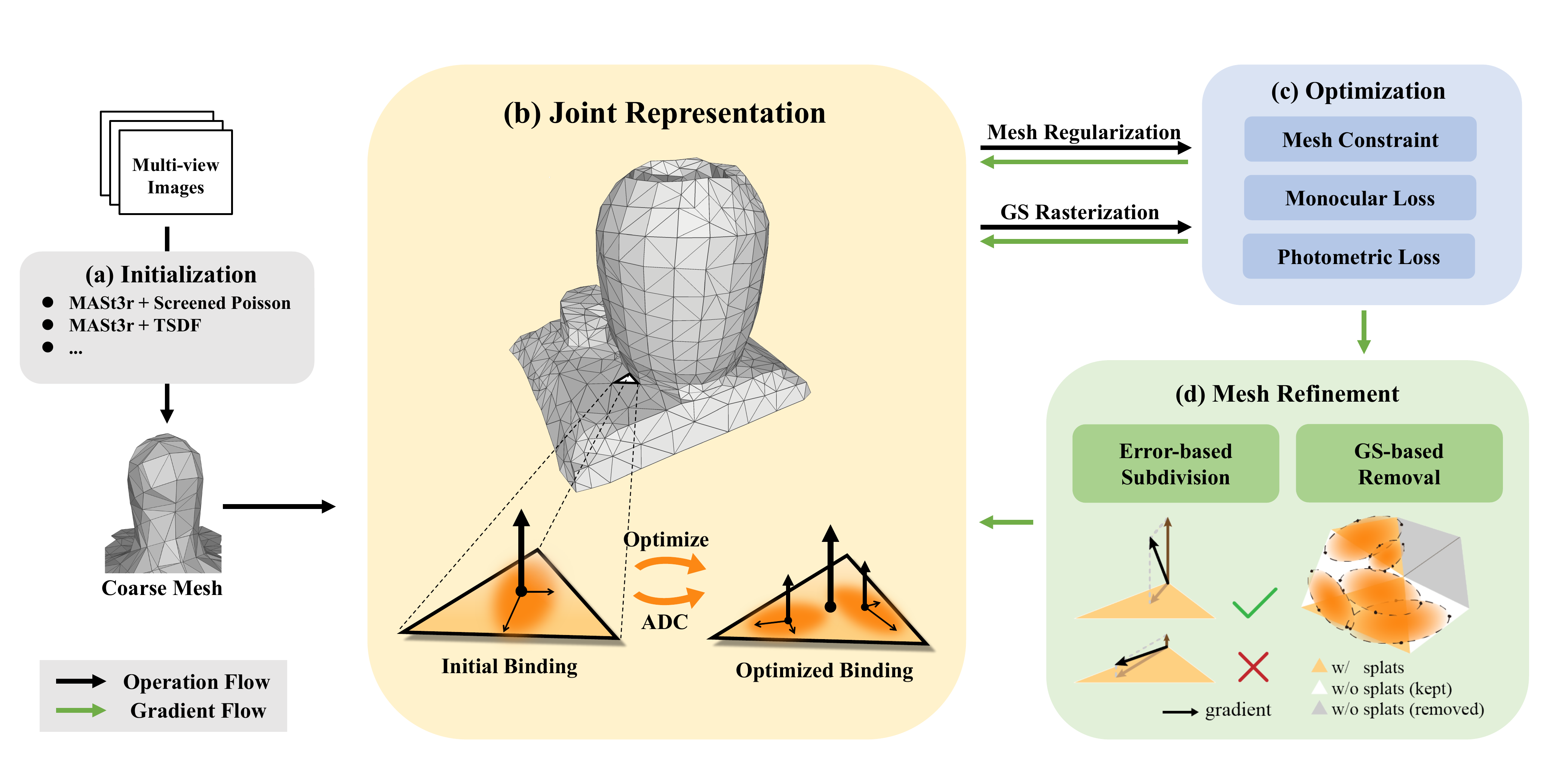}
  \caption{\textbf{An Overview of OMeGa}: We propose a novel end-to-end differentiable framework for both novel view synthesis and geometry reconstruction tasks. (a) Multi-view images are used to initialize a coarse mesh structure(\cref{sec:initial}). (b) We introduce a novel joint representation that associates 2D splats with mesh faces(\cref{sec:binding}). (c) The training process integrates mesh constraints and monocular geometry supervision to guide the optimization(\cref{sec:optimization}). (d) We further propose an iterative mesh refinement strategy to improve mesh quality(\cref{sec:refine}).}
  \label{fig:overview}
\end{figure*}

Given posed images, our goal is to simultaneously optimize 2D splats and a coarse mesh from scratch, to produce both high-fidelity rendered views and a high-quality surface reconstruction. Firstly, we introduce a joint representation of 2D splats and a triangle mesh (\cref{sec:binding}), ensuring that the splats both adhere closely to the surface geometry and retain sufficient flexibility. Then we will discuss the optimization objectives and geometric constraints used during the joint training process (\cref{sec:optimization}). To obtain a fine-grained and accurate reconstructed mesh, we carefully design a mesh refinement strategy to gradually subdivide and remove triangle faces during the optimization process (\cref{sec:refine}). Lastly, we will discuss our method to obtain the initial mesh through MASt3R\cite{leroy2024grounding} and Screened Poisson\cite{kazhdan2013screened} (\cref{sec:initial}). An overview of our framework is shown in \cref{fig:overview}.

\subsection{Joint Representation Design}
\label{sec:binding}
Vanilla 2DGS often misaligns with the true surface in texture-less regions due to insufficient geometric supervision, which yields inaccurate surface extraction. To mitigate this, we explicitly model spatial relations between neighboring splats and the mesh: a joint representation that flexibly binds 2DGS to mesh faces so that splat optimization benefits from mesh-derived geometric guidance.

Previous works\cite{qian2024gaussianavatars} typically allocate a fixed number of Gaussian splats uniformly to each mesh face, irrespective of geometric or textural complexity. Although some works\cite{waczynska2024games} allow splats to move within the surface, other attributes related to their representational capacity (e.g., rotation and scale) remain fixed and non-optimizable. However, we notice that geometry and texture exhibit inherently different distributions. For example, geometrically flat surfaces can display rich textures that require splats of various attributes for accurate representation. As shown in \cref{fig:binding}, we introduce a flexible binding that decouples geometric support from textural detail and adapts the number and placement of splats per face.

Specifically, our model contains two parameters types: (i) \emph{mesh} parameters which include learnable vertex positions \(\mathbf{V}\in\mathbb{R}^{N_v\times 3}\) and faces \(\mathbf{F}\in\mathbb{R}^{N_f\times 3}\); (ii) \emph{Gaussian Splatting} parameters used for rasterization. 

\noindent\textbf{Flexible Binding Strategy.} Vanilla 2DGS model consists of the following parameters: means, scales, rotations, colors, and opacities. The first three parameters characterize the spatial distribution, whereas the others describe the texture attributes. To make the splats align with the surface, spatial attributes in OMeGa are expressed mainly in terms of mesh parameters. Furthermore, to achieve stronger texture expressiveness, we preserve the texture attributes in the splats and introduce adaptive coefficients into the spatial attributes, as well as allowing adaptive density control proposed in \cite{kerbl20233d}.

Formally, we maintain a mapping from each splat to its supporting face during training:
\begin{equation}
    \mathcal{B}:\; \mathbf{G} \rightarrow \mathbf{F},\qquad \mathcal{B}(g_i)=f_j .
    \label{eq:binding}
\end{equation}

During optimization, both the splat set $\mathbf{G}$ and the mesh topology $\mathbf{F}$ may change; we update $\mathcal{B}$ with the following rules:
\begin{enumerate}
\item \textbf{Splat cloning and splitting.} When a Gaussian $g_i$ bound to $f_j$ is cloned or split, a new splat $g_{\text{new}}$ is created and initially bound to the same face $f_j$, inheriting $g_i$'s barycentric position with a small in-plane offset onto the face plane $\Pi_{f_j}$. We then insert $(g_{\text{new}},f_j)$ into $\mathcal{B}$.

\item \textbf{Splat removal.} If a splat is pruned by the adaptive density controller, its entry is removed from $\mathcal{B}$.

\item \textbf{Mesh subdivision.} When a face $f_j$ is subdivided into children $\{f_{j_k}\}$ (see \cref{sec:refine}),
each splat previously bound to $f_j$ is reassigned to the child whose domain contains its barycentric coordinates; per-splat attributes are preserved.

\item \textbf{Mesh Removal.} When a face $f_j$ is removed (see \cref{sec:refine}), all splats mapped to $f_j$ are identified and flagged for removal in subsequent steps.
\end{enumerate}

To allow splats to move freely on the face, the centers of splats are derived from the barycentric coordinates associated with each triangular mesh face. For each face with vertices denoted as \( v_1 \), \( v_2 \), and \( v_3 \), we parameterize the corresponding 2DGS mean \( \boldsymbol{m} \) as:
\begin{equation}
\boldsymbol{m} = \boldsymbol{u} \cdot v_1 + \boldsymbol{w} \cdot v_2 + \left( \mathit{1-\boldsymbol{u}-\boldsymbol{w}}\right)\cdot v_3
\end{equation}
where  $\boldsymbol{u},\boldsymbol{w} \in[0,1]$ and $ \boldsymbol{u}+\boldsymbol{w} \leq 1$. 

To prevent significant discrepancies between the Gaussian Splats' scales and the scales of the corresponding triangular faces, we impose upper and lower bounds for each Gaussian Splats' scale $\boldsymbol{s}$, which can be formulated as:
\begin{equation}
    \boldsymbol{s} = \boldsymbol{s}_{rel} \odot \boldsymbol{s}_0
\end{equation}
where $\boldsymbol{s}_{rel}=(s_{rel,1},s_{rel,2})\in [s_{min},s_{max}]\times[s_{min},s_{max}]$ represents the coefficient vector, $\boldsymbol{s}_0$ represents the predefined scale derived from the mesh faces, $\odot$ denotes element-wise multiplication. 

Furthermore, to model the orientation of Gaussian Splats on the face plane, we introduce an in-plane rotation parameter $\boldsymbol{R_{2d}}$, which can be transformed into world space via the following formula:
\begin{equation}
\boldsymbol{R} = \boldsymbol{R_s}\cdot \boldsymbol{R_{2d}}
\end{equation}
where $\boldsymbol{R}$ denotes the final rotation matrix used for rasterization and $\boldsymbol{R_s}$ is the rotation matrix of triangle faces. 

\noindent\textbf{Illumination handling.} Additionally, we notice that conventional spherical harmonic representation can not handle intense illumination changes, a common situation if the indoor scene contains multiple light sources. Following WildGaussians\cite{kulhanek2024wildgaussians}, we replace the spherical harmonics with a feature embedding $\boldsymbol{f}$ and trainable per-image embedding $\boldsymbol{e}$. Finally, we retain the opacity parameter $\boldsymbol{\alpha}$ consistent with the standard 2D Gaussian Splatting representation.

\subsection{Optimization}
\label{sec:optimization}
\noindent\textbf{Photometric Constraints.} Following 3DGS\cite{kerbl20233d}, we utilize the combination of $\mathcal{L}_1$ and $\mathcal{L}_{D-SSIM}$ between the rendered image and the ground truth image, which is expressed as follows:
\begin{equation}
    \mathcal{L}_{img} = \lambda_{1}\mathcal{L}_1 + (1-\lambda_{1})\mathcal{L}_{D-SSIM}
\end{equation}
where $\lambda_1$ controls the weight of $\mathcal{L}_1$ loss.

\noindent\textbf{Mesh Constraints.} We observe that a straightforward vertex optimization often yields a disordered mesh topology during training, and once such topological errors appear, subsequent steps struggle to revert them to a coherent structure. Drawing inspiration from previous works on mesh optimization\cite{liu2025dynamicgaussiansmeshconsistent,choi2024ltm}, we utilize Laplacian smooth loss\cite{desbrun1999implicit,nealen2006laplacian} and normal consistency loss as regularization terms to reduce the occurrence of mesh distortion (\eg intersecting triangles) during the optimization process. The Laplacian smooth loss is 
\begin{equation}
\mathcal{L}_{lap} = \frac{1}{|\boldsymbol{V}|} \sum_{i \in \boldsymbol{V}} \left\| v_i - \frac{1}{|\mathcal{N}(i)|} \sum_{j \in \mathcal{N}(i)} v_j \right\|^2
\end{equation} 
where $\mathcal{N}(i)$ represents the set of neighbor vertex of given vertex $v_i$. The Laplacian smooth loss minimizes the distance between each vertex and the average location of its neighbors. The normal consistency loss is 
\begin{equation}
\mathcal{L}_{nc} = \frac{1}{|\mathcal{N}_f|} \sum_{(i, j) \in \mathcal{N}_f} \left(1 - \boldsymbol{n}_{i} \cdot \boldsymbol{n}_{j} \right)
\end{equation}
where $\mathcal{N}_f$ denotes the set of all pairs of neighboring faces in the mesh, $\boldsymbol{n}_i$ and $\boldsymbol{n}_j$ represent the surface normal of adjacent mesh faces. The normal consistency constraint encourages a smoother surface during the optimization process and reduces the occurrence of mesh faces flipping. 

While the terms themselves are conventional, OMeGa's novelty lies in embedding them within a joint optimization over an explicit mesh and Gaussian splats, allowing these constraints to influence Gaussian splats learning via the binding strategy. This turns the Laplacian and normal-consistency losses from post-hoc smoothers into geometric priors that substantially constrain the optimization of mesh vertices and improve surface quality.

\noindent\textbf{Monocular Geometry Constraints.} Previous works\cite{turkulainen2024dn,zhang20242dgs,wu2024surface} on indoor scene reconstruction have explored methods to improve reconstruction results using foundational models\cite{bochkovskii2024depth,bae2021estimating,bhat2023zoedepth,eftekhar2021omnidata}. We propose to utilize normal cues $\hat{\boldsymbol{N}}$ from Stable Normal\cite{ye2024stablenormal} as a monocular geometry supervision during the optimization process, which can be expressed as follows:

\begin{equation}
\mathcal{L}_{n} = \sum \max \left\{ \lVert \boldsymbol{N} - \hat{\boldsymbol{N}} \rVert - m, 0 \right\}
\end{equation}
where $m$ is a small margin to handle the inherent normal inconsistency across different views produced by monocular foundational models.

\noindent\textbf{Training Objective.} In summary, we minimize the following loss function during the optimization process:
\begin{equation}
\mathcal L = \mathcal{L}_{img} + \lambda_{lap} \mathcal{L}_{lap} + \lambda_{nc} \mathcal{L}_{nc} + \lambda_n \mathcal{L}_n
\end{equation}
where $\lambda_{lap}$, $\lambda_{nc}$, and $\lambda_n$ control the weight of Laplacian smooth loss, normal consistency loss, and monocular normal loss, respectively.

\subsection{Mesh Refinement Strategy}
\label{sec:refine}
Though the proposed joint representation significantly enhances the reconstructed mesh results, we observe that the resulting mesh remains suboptimal. Firstly, the density of mesh faces remains nearly the same throughout the whole scene, while high-frequency details should be fit with more triangle faces, and low-frequency areas can be fit with fewer triangle faces. Therefore, we propose an error-based coarse-to-fine mesh subdivision strategy to adaptively densify triangle faces, resulting in a more detailed reconstructed mesh. Secondly, we notice that the initial mesh often includes extraneous triangle faces not present in the actual scene, leading to inconsistencies between the distributions of Gaussian Splats and mesh triangles. To resolve this issue, we present a mesh faces removal strategy to maintain the consistency between 2D Gaussian Splats and the mesh.

\noindent\textbf{Error-based Mesh Subdivision.}  During the optimization process, we notice that the gradients of vertices are robust indicators for high-geometry frequency areas. Inspired by the coarse-to-fine mesh refinement strategy proposed in NeRF2Mesh\cite{tang2022nerf2mesh}, we design an iterative mesh subdivision method, which is based on the gradients of mesh vertices. Specifically, we project the gradients of vertices into the normal direction of the faces they belong to as follows:
\begin{equation}
    e(f_j) = \sum_{v_i \in f_j} \left\|(\nabla_{v_i}\mathcal{L}) \cdot \boldsymbol{n}_{f_j}\right\|
\end{equation}
where $\nabla_{v_i}\mathcal{L}$ represents the gradient of vertices $v_i$ and $\boldsymbol{n}_{f_i}$ is the normal of face $f_i$. Subsequently, we perform one iteration of the midpoint subdivision algorithm on the mesh faces corresponding to the top-k largest accumulated gradient magnitudes $e(f_j)$. After conducting the subdivision algorithm, we initialize the parameters of newly-generated Gaussian Splats by transferring parameters from their nearest previously optimized Gaussian Splats, thus preserving the learned features and accelerating convergence. In this way, high-frequency geometric regions (\eg, intricate indoor details) are preferentially split into more faces to better fit the shape, whereas low-frequency areas (\eg, floors and walls) remain low-complexity. Leveraging this normal-projected gradient signal, our refinement further decouples geometry optimization from appearance modeling.

\noindent\textbf{GS-based Mesh Faces Removal.} The initial coarse mesh contains extraneous geometry that is inconsistent with the observed scene. Although splats binding with these faces are removed by the ADC optimization strategy, the mesh faces persist in the scene during optimization, which in turn degrades reconstruction accuracy. To this end, we propose an approach to gradually remove irrelevant mesh faces by analyzing the spatial distribution of Gaussian splats and mesh faces. Specifically, we sample points uniformly at the endpoints of the major and minor axes of the 2D splats, yielding a set of 3D points that represent the location distribution of Gaussian splats, denoted as $\boldsymbol{V'}$. Then, we calculate the Chamfer distance between the vertices of the original mesh faces $\boldsymbol{V}$ and the derived surfel vertices $\boldsymbol{V'}$. Mesh faces whose Chamfer distance exceeds a predefined removal threshold $\tau_{remove}$ are then eliminated during the training process.

\subsection{Initial Mesh Reconstruction}
\label{sec:initial}
In recent works\cite{waczynska2024games, lin2025directlearningmeshappearance}, a pre-trained Gaussian Splatting model is often required to extract a detailed mesh for initialization. In contrast, with our proposed optimization and refinement strategy, a coarse initial mesh suffices to achieve high-quality mesh reconstruction through training. Therefore, a variety of simple and training-free approaches can be employed in our pipeline for initialization — for example, using image-to-3D models to reconstruct a dense but potentially noisy point cloud from images, followed by Poisson surface reconstruction to generate a coarse mesh.

Although MASt3R\cite{leroy2024grounding} and Screened Poisson\cite{kazhdan2013screened} are adopted for mesh initialization in our experiments, we emphasize that our framework is agnostic to a specific reconstruction method. Any approach capable of producing a coarse mesh can be employed for the subsequent mesh generation and Gaussian Splats initialization.
\newcommand{\First}{\cellcolor[HTML]{FFB2B2}}
\newcommand{\Second}{\cellcolor[HTML]{FFD8B2}}
\newcommand{\Third}{\cellcolor[HTML]{FFFFB2}}

\begin{table*}[t]
    \centering
    \resizebox{\textwidth}{!}
    {
    \begin{tabular}{l|ccc|ccccc|ccccc|ccccc}
        \toprule
        \multirow{2}{*}{Method} & & & & \multicolumn{5}{c|}{MuSHRoom \cite{ren2023mushroom}} &  \multicolumn{5}{c|}{ScanNet \cite{dai2017scannet}} & \multicolumn{5}{c}{ScanNet++ \cite{yeshwanth2023scannet++}} \\
         & N & E & M & Acc.↓ & Comp.↓ & C-$L_1$↓ & NC↑ & F1↑ & Acc.↓ & Comp.↓ & C-$L_1$↓ & NC↑ & F1↑ & Acc.↓ & Comp.↓ & C-$L_1$↓ & NC↑ & F1↑ \\ 

\midrule
2DGS\cite{huang20242d}$^\dagger$ & & & & 0.0862 & 0.0708 & 0.0785 & 0.7533 & 0.5575 & 0.0860 & 0.0953 & 0.0907 & 0.7174 & 0.4859 & 0.0974 & 0.0677 & 0.0825 & 0.7586 & 0.4807 \\
GOF\cite{yu2024gaussian}$^\dagger$ & & & & 0.1819 & 0.0730 & 0.1275 & 0.6501 & 0.4073 & 0.1369 & 0.0839 & 0.1104 & 0.6130 & 0.4905 & 0.1136 & 0.0656 & 0.0896 & 0.7184 & 0.4486\\
PGSR\cite{chen2024pgsr}$^\dagger$ & & & & 0.0971 & 0.1004 & 0.0988 & 0.7362 & 0.5060 & 0.0558 & 0.0693 & 0.0626	& 0.7537 & 0.6153 & 0.1052 & 0.0840 & 0.0946 & 0.7828 & 0.4044\\
\midrule
GSRec\cite{wu2024surface} & \checkmark & & & 0.0612 & 0.0630 & 0.0621 & 0.7053 & 0.6450 & 0.0748 & 0.0718 & 0.0733 & 0.6544 & 0.5462 & 0.1135 & 0.1654 & 0.1395 & 0.6913 & 0.3691\\
2DGS\cite{huang20242d}$^\dagger$ & \checkmark & & & 0.0887 & 0.0731 & 0.0809 & 0.7707 & 0.5727 & 0.0804 & 0.0801 & 0.0802 & 0.7676 & 0.5563 & 0.0938 & 0.0669 & 0.0803 & 0.8037 & 0.5373\\
2DGS\cite{huang20242d}$^\dagger$ & \checkmark & \checkmark & & 0.0817 & 0.0659 & 0.0738 & \Third 0.7902 & 0.6145 & 0.0706 & 0.0710 & 0.0708 & \Third 0.7853 & 0.5900 & 0.0599 & 0.0565 & 0.0582 & \Third 0.8543 & 0.6315\\
\midrule
SuGaR\cite{guedon2024sugar} & & & \checkmark & 0.0676 & 0.0556 & 0.0616 & 0.6957 & 0.6268 & 0.0987 & 0.0975 & 0.0981 & 0.6641 & 0.5058 & 0.1949 & 0.1098 & 0.1524 & 0.6567 & 0.3858 \\
GaMeS\cite{waczynska2024games}$^\dagger$ & & & \checkmark & \Third 0.0585 & \First 0.0379 & \Third 0.0482 & 0.7088 & \Third 0.7211 & \Third 0.0458 & \First 0.0431 & \First 0.0445 & 0.6307 & \Third 0.7282 & \Third 0.0546 & \First 0.0287 & \Third 0.0416 & 0.6827 & \Third 0.7675\\
\midrule
\textbf{Ours}$^\dagger$ & \checkmark & & \checkmark & \Second 0.0405 & \Third 0.0441 & \Second 0.0423 & \Second 0.8571 & \Second 0.7606 & \Second 0.0391  & \Third 0.0528 & \Third 0.0470 & \Second 0.8343 & \Second 0.7286 & \Second 0.0284 & \Second 0.0378 & \First 0.0331 & \Second 0.9120 & \Second 0.8559\\
\textbf{Ours}$^\dagger$ & \checkmark & \checkmark & \checkmark & \First 0.0398 & \Second 0.0431 & \First 0.0414 & \First 0.8586 & \First 0.7668 & \First 0.0383 & \Second 0.0525 & \Second 0.0454 & \First 0.8364 & \First 0.7317 & \First 0.0264 & \Third 0.0399 & \Second 0.0331 & \First 0.9144 & \First 0.8631\\
        \midrule
        \textit{Improvement} &  &  &  & 
        +53.8\% & 
        +39.1\% & 
        +47.3\% & 
        +14.0\% & 
        +37.5\% & 
        +55.5\% & 
        +44.9\% & 
        +49.9\% & 
        +16.6\% & 
        +50.6\% & 
        +72.9\% & 
        +41.1\% & 
        +59.9\% & 
        +20.5\% & 
        +79.6\% \\
        \bottomrule
    \end{tabular}
    } 
    \caption{\textbf{Quantitative reconstruction comparison on MuSHRoom, ScanNet and ScanNet++ dataset.} Averaged results are reported over 5 scenes, 8 scenes, and 2 scenes, respectively. $^\dagger$ indicates initialized with MASt3R priors. The Improvement row is computed relative to 2DGS$^\dagger$. Columns N, E, and M indicate whether monocular normal supervision, feature embeddings, and a mesh representation are used, respectively. The best result is highlighted in 
\colorbox[HTML]{FFB2B2}{red}, the second best result is highlighted in  \colorbox[HTML]{FFD8B2}{orange}, and the third best result is highlighted in \colorbox[HTML]{FFFFB2}{yellow}.}
    \label{tab:reconstruction_quality}
\end{table*}

\begingroup
\setlength{\tabcolsep}{12pt}
\begin{table*}[t]
    \centering
    \resizebox{\textwidth}{!}
    {
    \begin{tabular}{l|ccc|ccc|ccc|ccc}
        \toprule
        \multirow{2}{*}{Method} & & & & \multicolumn{3}{c|}{MuSHRoom \cite{ren2023mushroom}} &  \multicolumn{3}{c|}{ScanNet \cite{dai2017scannet}} & \multicolumn{3}{c}{ScanNet++ \cite{yeshwanth2023scannet++}} \\
         & N & E & M & PSNR ↑ & SSIM ↑ & LPIPS & PSNR ↑ & SSIM ↑ & LPIPS & PSNR ↑ & SSIM ↑ & LPIPS \\ 

\midrule
2DGS\cite{huang20242d}$^\dagger$ & & & & \Third 24.03 & \Third 0.834 & \Second 0.232 & 23.39 & 0.827 & 0.448 & \Third 23.82 & \Second 0.913 & \Third 0.242 \\
GOF\cite{yu2024gaussian}$^\dagger$ & & & & 21.61 & 0.789 & 0.279 & 22.35 & 0.810 & \Third 0.370 & 21.66 & 0.894 & \Third 0.242 \\
PGSR\cite{chen2024pgsr}$^\dagger$ & & & & 22.95 & \First 0.893 & 0.271 & \Third 23.85 & \Second 0.832 & \First 0.360 & 22.95 & 0.893 & 0.271 \\
\midrule
GSRec\cite{wu2024surface} & \checkmark & & & 22.22 & 0.784 & 0.409 & 22.82 & 0.813 & 0.416 & 21.67 & 0.883 & 0.399 \\
2DGS\cite{huang20242d}$^\dagger$ & \checkmark & & & 23.67 & 0.827 & 0.248 & 23.47 & 0.828 & 0.440 & 23.20 & 0.899 & 0.286 \\
2DGS\cite{huang20242d}$^\dagger$ & \checkmark & \checkmark & & \First 24.53 & \Second 0.836 & \First 0.225 & \First 24.36 & \First 0.837 & 0.419 & \Second 26.42 & \Third 0.912 & 0.249 \\
\midrule
SuGaR\cite{guedon2024sugar} & & & \checkmark & 23.44 & 0.817 & 0.261 & 23.64 & \Third 0.829 & \Second 0.363 & 22.04 & 0.891 & 0.261\\
GaMeS\cite{waczynska2024games}$^\dagger$ & & & \checkmark & 22.93 & 0.800 & 0.287 & 23.69 & 0.828 & 0.376 & 22.82 & 0.893 & 0.255\\
\midrule
\textbf{Ours}$^\dagger$ & \checkmark & & \checkmark & 23.22 & 0.825 & 0.242 & 22.75 & 0.806 & 0.416 & 21.95 & 0.900 & 0.250\\
\textbf{Ours}$^\dagger$ & \checkmark & \checkmark & \checkmark & \Second 24.05 & 0.822 & \Third 0.240 & \Second 24.29 & 0.828 & 0.398 & \First 26.62 & \First 0.916 & \First 0.205\\
        \bottomrule
    \end{tabular}
    } 
    \caption{\textbf{Quantitative rendering comparison on MuSHRoom, ScanNet and ScanNet++ dataset.} Averaged results are reported over 5 scenes, 8 scenes and 2 scenes, respectively. $^\dagger$ indicates initailized with MASt3R priors. Columns N, E, and M indicate whether monocular normal supervision, feature embeddings, and a mesh representation are used, respectively. The best result is highlighted in 
\colorbox[HTML]{FFB2B2}{red}, the second best result is highlighted in  \colorbox[HTML]{FFD8B2}{orange}, and the third best result is highlighted in \colorbox[HTML]{FFFFB2}{yellow}.}
    \label{tab:rendering_quality}
\end{table*}
\endgroup

\section{Experiment}
In this section, we introduce our experiment settings in \cref{sec:experimental_settings} and primary experiment results in \cref{sec:results}. Further ablation studies and analysis are conducted in \cref{sec:ablation_study}.
\subsection{Experimental Settings}
\label{sec:experimental_settings}

\noindent\textbf{Implementation Details.} The implementation of our approach is mainly based on the PyTorch\cite{paszke2019pytorch} and gsplat\cite{ye2024gsplatopensourcelibrarygaussian} framework and experimented in a single RTX 4090 GPU. All hyperparameters and mesh-extraction details are provided in \cref{appendix:implementation}.

\noindent\textbf{Datasets.} We conduct a comprehensive evaluation of our model and other baseline models using real-world datasets. We utilize 5 scenes from the MuSHRoom\cite{ren2023mushroom} dataset, 8 scenes from the ScanNet\cite{dai2017scannet} dataset, and 2 scenes from the ScanNet++\cite{yeshwanth2023scannet++} dataset.

\noindent\textbf{Metrics.} For novel view synthesis tasks, we evaluate rendered images using PSNR, SSIM\cite{wang2004image}, and LPIPS\cite{zhang2018unreasonable} metrics consistent with previous works\cite{kerbl20233d,huang20242d} on neural rendering. For reconstructed meshes, we report results based on five metrics: Accuracy (Acc.), Completion (Comp.), Chamfer-$L_1$ distance (C-$L_1$), Normal Consistency (NC), and F-scores following previous studies\cite{ren2024ags}. We also report additional evaluation metrics, such as runtime and GPU memory consumption, in \cref{appendix:metrics}.

\noindent\textbf{Baselines.} We compare our approach with state-of-the-art Gaussian Splatting methods for scene reconstruction. The baseline models include: GOF\cite{yu2024gaussian}, SuGaR\cite{guedon2024sugar}, PGSR\cite{chen2024pgsr}, GSRec\cite{wu2024surface}, 2DGS\cite{huang20242d} and GaMeS\cite{waczynska2024games}. For fair comparison, we initialize the baseline models using dense point clouds/meshes derived from MASt3R. We also report 2DGS with the same normal supervision and feature embedding used in OMeGa to ensure fairness in comparing reconstruction and rendering results.

\label{sec:results}
\begin{figure*}[htb]
  \includegraphics[width=0.97\linewidth]{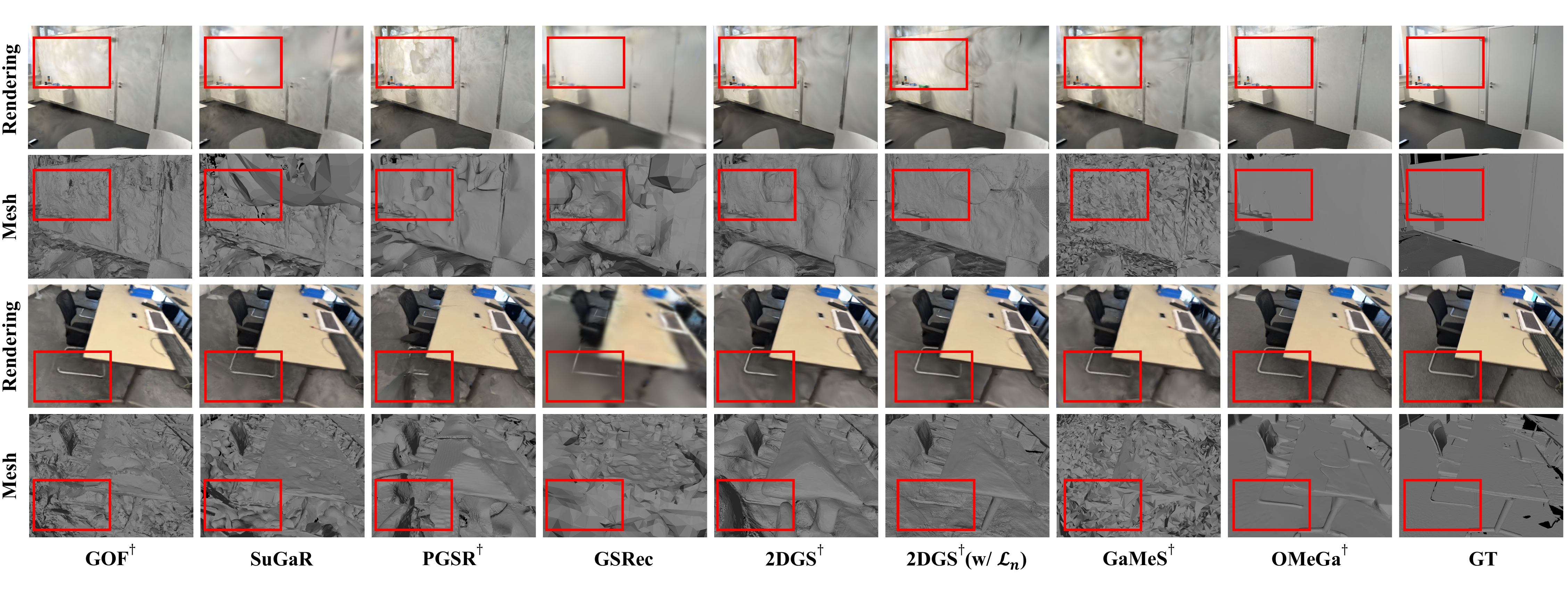}
  \caption{\textbf{Qualitative results on the ScanNet++ dataset}: In each scene, our method yields the highest-quality mesh reconstructions, closely matching the ground-truth geometry, especially in texture-less areas. $^\dagger$ indicates initailized with MASt3R priors.}
  \label{fig:qualitative_analysis}
\end{figure*}
\subsection{Experimental Results}
\noindent\textbf{Geometry reconstruction results.} We report the quality of reconstructed mesh in \cref{tab:reconstruction_quality} on real-world indoor room scenes in the MuSHRoom\cite{ren2023mushroom} dataset, ScanNet\cite{dai2017scannet} dataset and ScanNet++\cite{yeshwanth2023scannet++} dataset. Our approach achieves state-of-the-art reconstruction results in comparison with previous Gaussian Splatting-based methods. Relative to the 2DGS$^\dagger$ baseline with TSDF post-hoc extraction, OMeGa achieves F-score gains of 37.5\%, 50.6\%, and 79.6\% on MuSHRoom, ScanNet, and ScanNet++, respectively.
Although many baselines initialize from MASt3R priors, the lack of explicit mesh structure constraints causes them to forget geometry priors (e.g., that texture-less regions should be flat) during optimization. In contrast, we convert the dense point cloud into a coarse mesh, explicitly restoring these priors in the initial surface, and then refine the model under mesh-guided optimization. 
Additionally, the mesh refinement strategy and optimization constraints also contribute to the reconstruction results, which will be further discussed in \cref{sec:ablation_study}.

\begin{figure}[htb]
  \centering
  \begin{subfigure}[b]{0.325\linewidth}
    \centering
    \includegraphics[width=\linewidth]{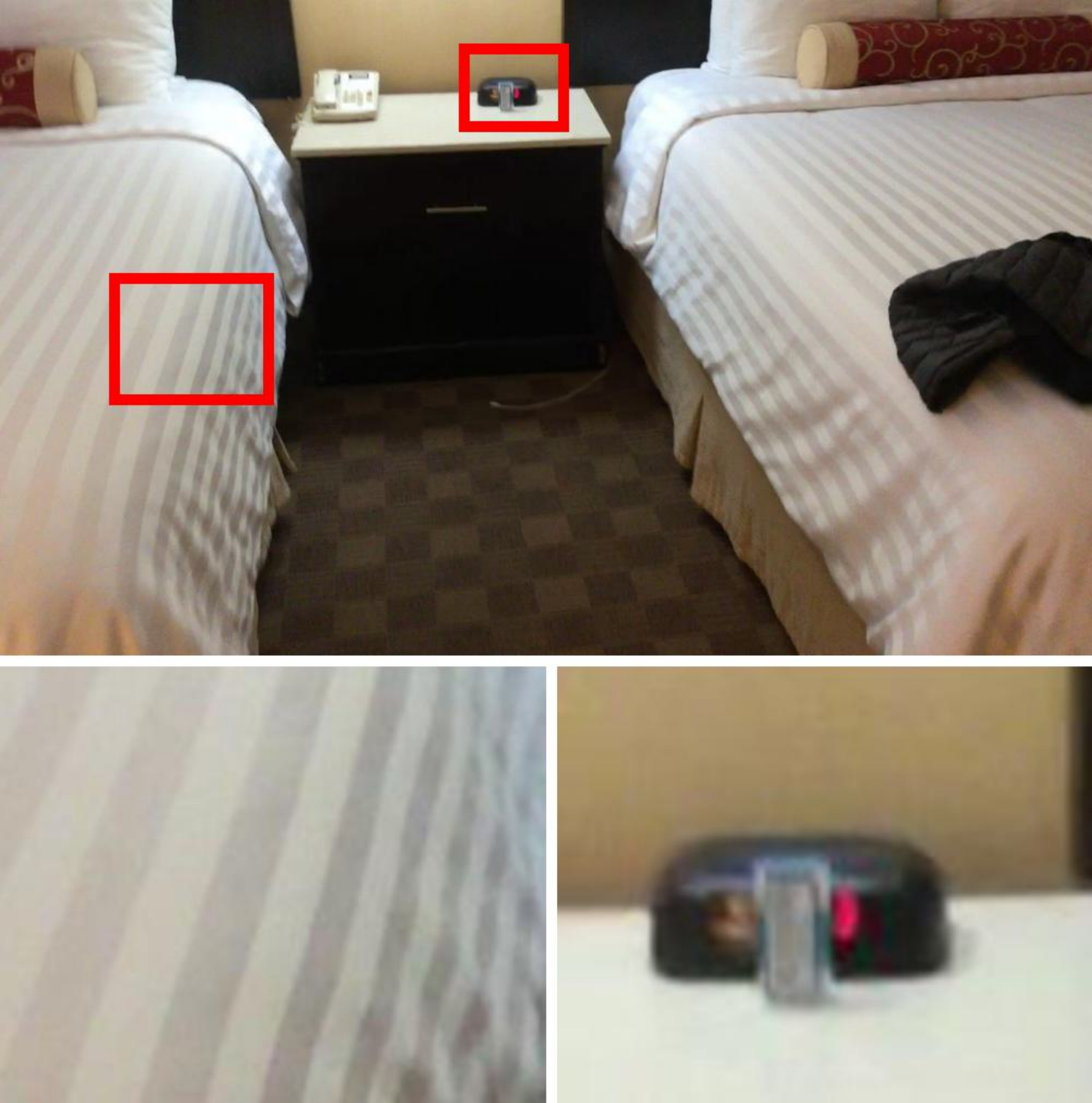}
    \caption{reference GT view}
    \label{fig:render_gt}
  \end{subfigure}
  \hfill
  \begin{subfigure}[b]{0.325\linewidth}
    \centering
    \includegraphics[width=\linewidth]{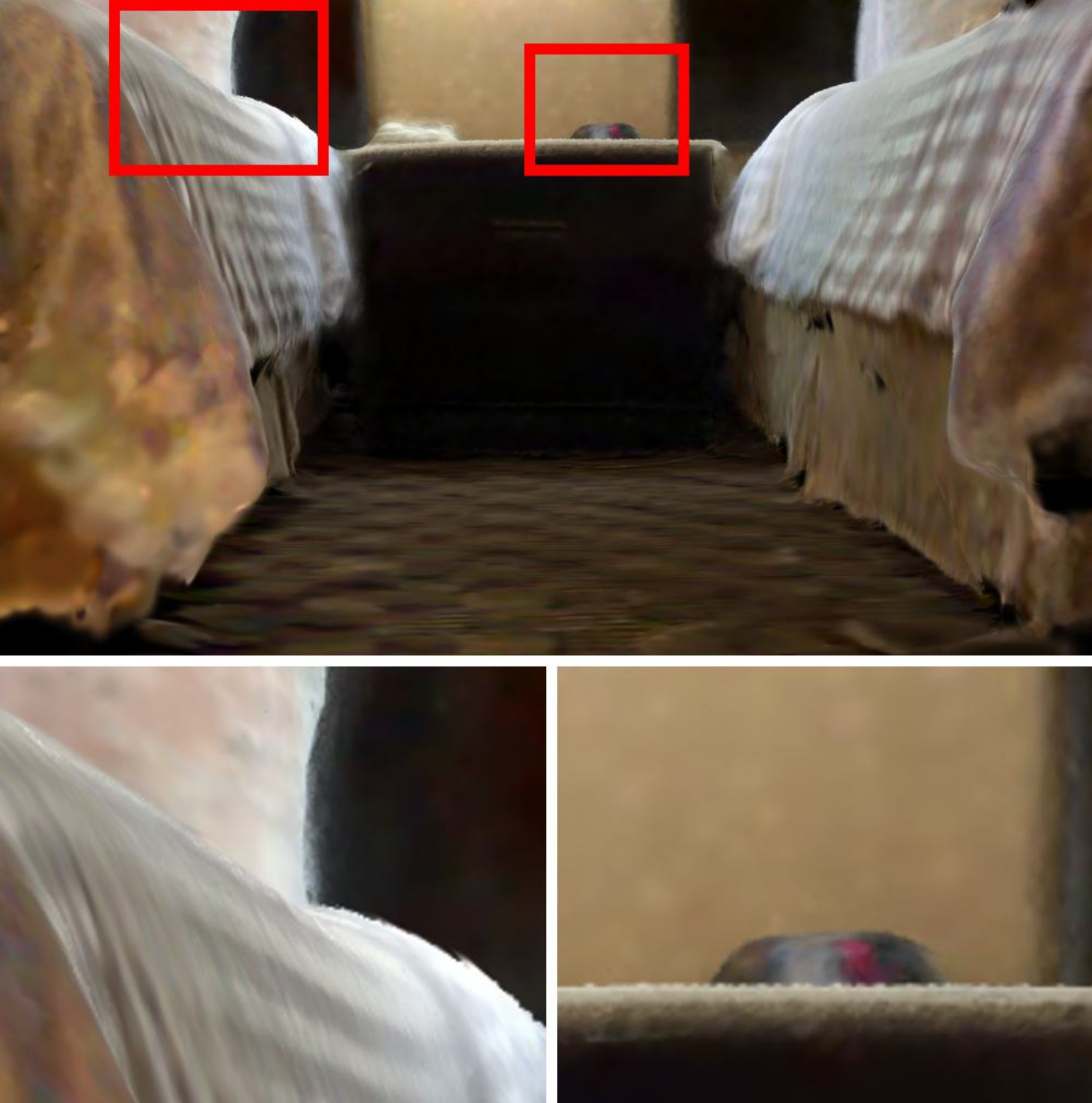}
    \caption{Ours}
    \label{fig:render_ours}
  \end{subfigure}
  \hfill
  \begin{subfigure}[b]{0.325\linewidth}
    \centering
    \includegraphics[width=\linewidth]{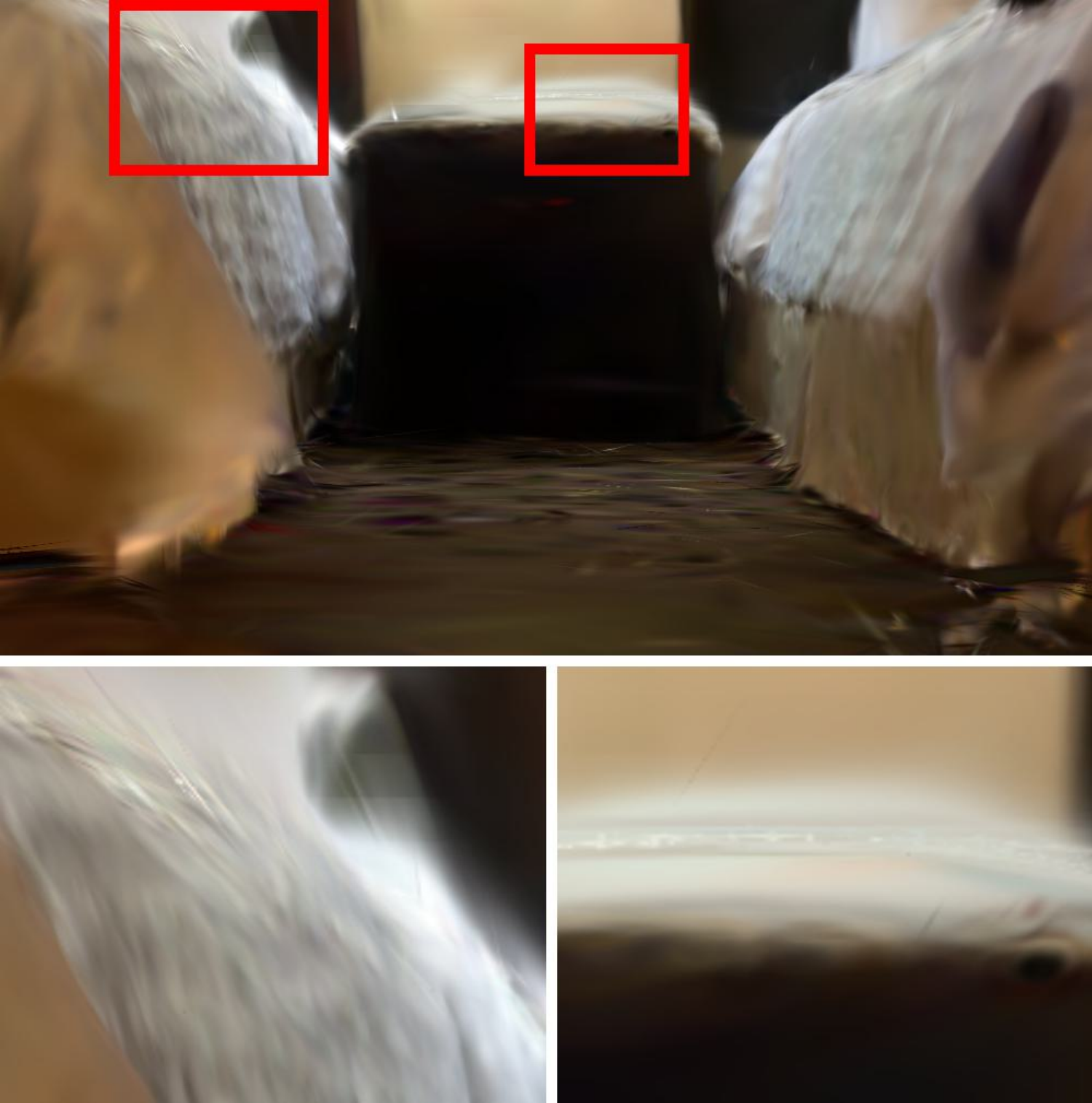}
    \caption{2DGS$^\dagger$(w/ $\mathcal{L}_n$)}
    \label{fig:render_2dgs}
  \end{subfigure}
  \caption{\textbf{Qualitative results of out-of-domain view's rendering.} On the out-of-domain viewpoint, OMeGa renders finer details than the 2DGS baseline.}
  \label{fig:render_new_view}
\end{figure}

\noindent\textbf{Novel view synthesis results.} Quantitative results for novel views rendering quality are reported in \cref{tab:rendering_quality}. On MuSHRoom and ScanNet datasets, OMeGa is comparable to 2DGS$^{\dagger}$ and 3DGS on the majority of scenes and surpasses most other baselines. We attribute this outcome to the evaluation process, wherein test views are sampled along the same capture trajectory as training views with minimal pose variation. This setup can inadvertently reward overfitting to appearance, yielding high metrics even when the underlying reconstructed geometry is inaccurate. Because OMeGa achieves accurate geometry reconstruction, it avoids such overfitting and therefore yields comparable but not superior rendering results on these in-distribution views. To assess generalization, we include an out-of-domain viewpoint absent from training (See \cref{fig:render_new_view}); OMeGa preserves detailed texture(\eg, floor, tabletop, and bedsheets), whereas baselines exhibit pronounced blur. This indicates that OMeGa trades a small amount of in-trajectory metrics for markedly better robustness under view shifts. On the ScanNet++ dataset, which exhibits pronounced illumination variation, OMeGa achieves the best rendering results among the compared methods. Methods without explicit illumination handling often overfit the illumination changes from different views by stacking Gaussian splats, which not only harm the geometry but also have limited robustness in novel views. In contrast, the use of feature embeddings provides robust rendering results based on the accurate geometry reconstruction brought by OMeGa. We include an additional discussion of illumination effects in \cref{appendix:illumination}.

\noindent\textbf{Qualitative analysis.} We present a qualitative analysis by visualizing reconstructed meshes and novel-view renderings in \cref{fig:qualitative_analysis}. Overall, our method consistently delivers smoother and more accurate geometry in texture-less regions (\eg walls and floors), which are challenging for previous approaches; compared with baselines, our results exhibit fewer irregularities and reduced depth discontinuities. Although GaMeS reports comparatively higher F-scores on some geometry metrics, its meshes frequently show topological instabilities (e.g., folded faces and self-intersections), whereas OMeGa remains largely topologically coherent, highlighting the benefit of incorporating mesh constraints as geometric priors during optimization. Adding monocular normal supervision further improves geometric fidelity in 2DGS$^\dagger$ (w/ $\mathcal{L}_n$) relative to the model without normal supervision, but normal supervision alone is insufficient to fully resolve texture-less areas, where depth discontinuities still exist in the reconstructed mesh. 
Collectively, these observations highlight the robustness of our approach for challenging indoor scene reconstruction.

\subsection{Ablations}
\label{sec:ablation_study}
We conduct ablation studies to evaluate the importance of key designs in our approach. The overall quantitative ablation results on the ScanNet++ dataset are shown in \cref{tab:ab_mesh}. 

\begin{figure}[htb]
  \centering
  \begin{subfigure}[b]{0.32\linewidth}
    \centering
    \includegraphics[width=\linewidth]{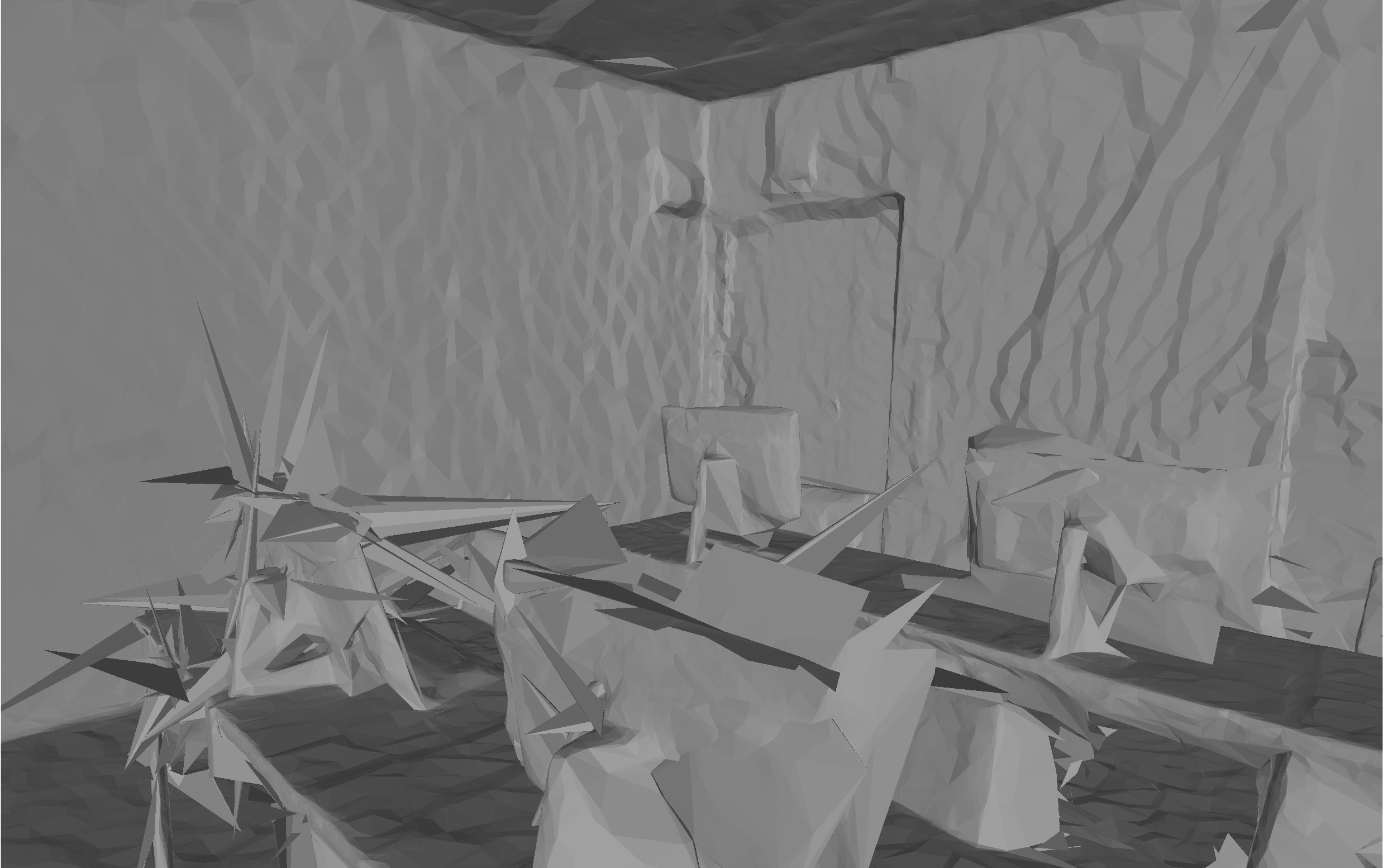}
    \caption{w/o $\mathcal{L}_{lap}$}
    \label{ablation_mesh_constraint_subfig_a}
  \end{subfigure}
  \hfill
  \begin{subfigure}[b]{0.32\linewidth}
    \centering
    \includegraphics[width=\linewidth]{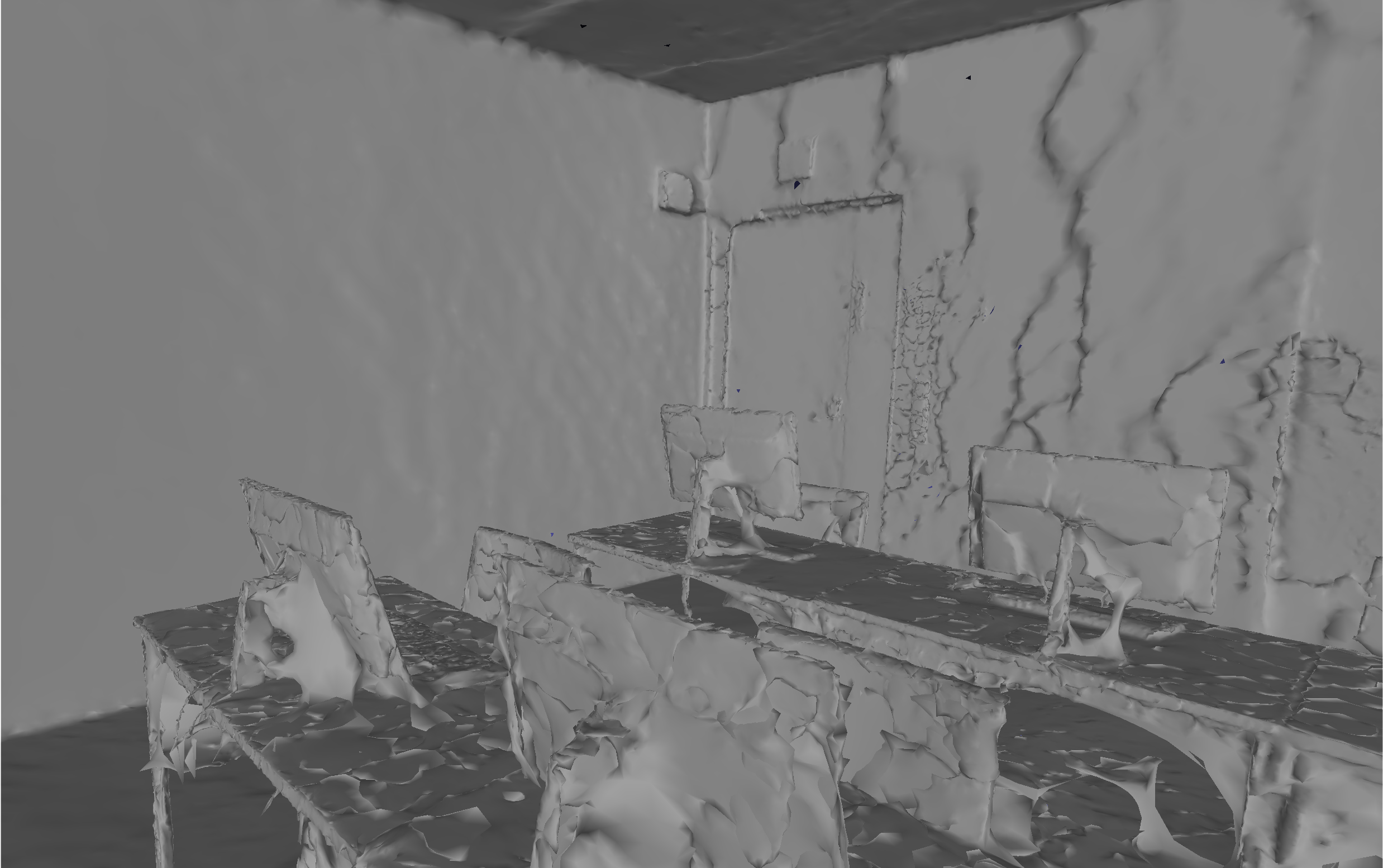}
    \caption{w/o $\mathcal{L}_{nc}$}
    \label{fig:ablation_mesh_constraint_subfig_b}
  \end{subfigure}
  \hfill
  \begin{subfigure}[b]{0.32\linewidth}
    \centering
    \includegraphics[width=\linewidth]{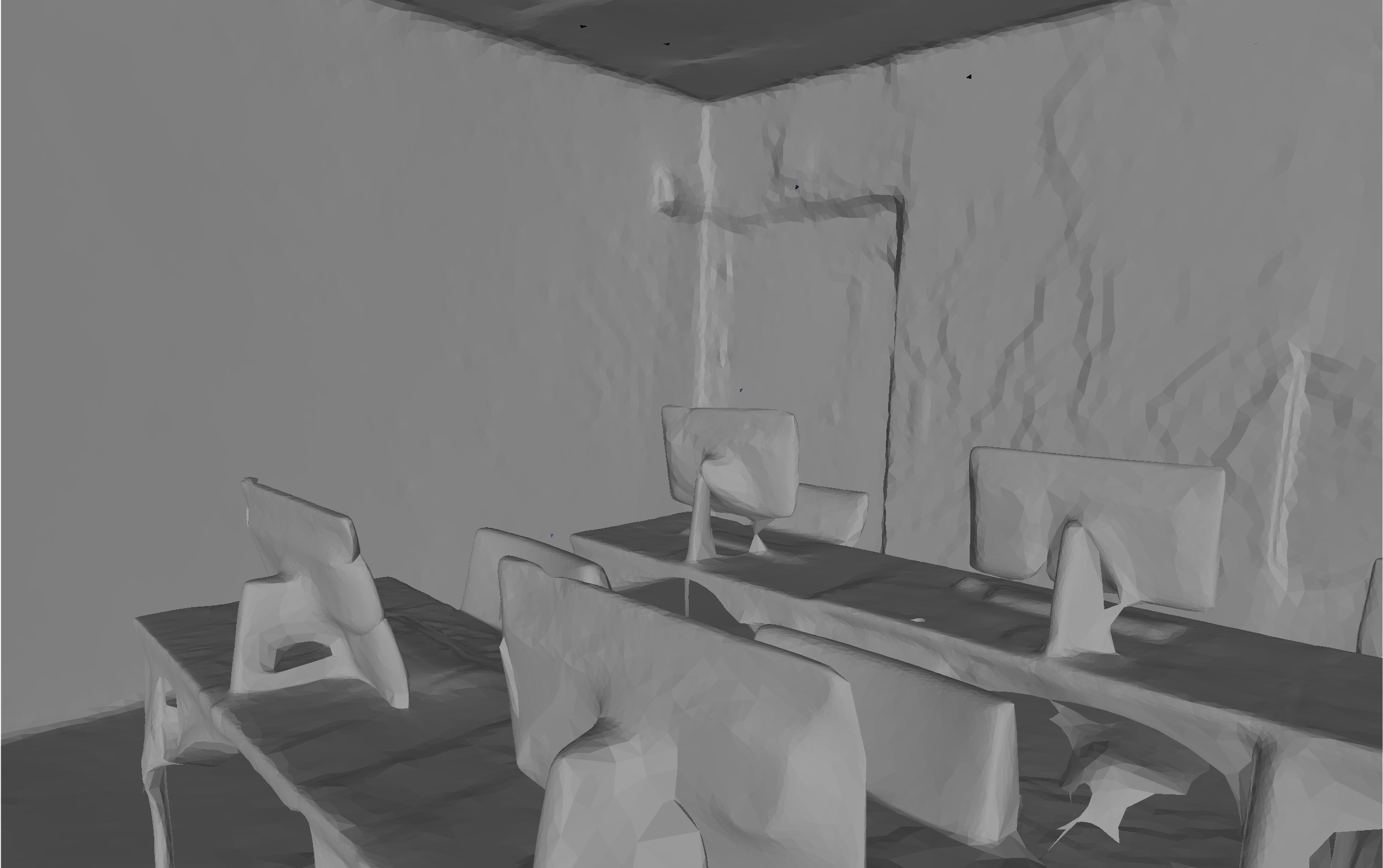}
    \caption{Full model}
    \label{ablation_mesh_constraint_subfig_c}
  \end{subfigure}
  \caption{\textbf{Qualitative results from ablation studies on mesh constraints.} The regularization improves mesh quality by producing smoother and more regular surfaces that better conform to the true geometry.}
  \label{fig:ablation_mesh_constraints}
\end{figure}

\noindent\textbf{Optimization Constraints.}
We evaluate the effects of mesh constraints and monocular normal supervision on reconstruction. As summarized in \cref{tab:ab_mesh}, the full model achieves the best reconstruction F\mbox{-}score, indicating the strongest overall surface quality.
Removing $\mathcal{L}_{n}$ produces a geometry drop among the losses. This confirms that $\mathcal{L}_{n}$ supplies orientation cues that photometric terms alone cannot resolve in texture-less regions. However, its performance still surpasses most baselines.
While removing Laplacian smoothness or normal consistency can slightly improve certain metrics, the resulting meshes exhibit pronounced self\mbox{-}intersections and surface irregularities, reflecting degraded geometric reconstruction during optimization.
Qualitative results on the MuSHRoom dataset (\cref{fig:ablation_mesh_constraints}) corroborate these observations: removing $\mathcal{L}_{lap}$ introduces needle-like spikes and irregular mesh faces, while training without $\mathcal{L}_{nc}$ causes face overlaps and flips. These artifacts substantially hinder the downstream use of the mesh. Collectively, Laplacian smoothing and normal consistency help produce smooth, topologically correct meshes, albeit with minor trade\mbox{-}offs in a few metrics. We therefore retain these constraints in practice to ensure stable geometry.

\begin{table}[H]
    \centering
     \resizebox{\columnwidth}{!}{%
    \begin{tabular}
    {lcccc>{\columncolor[gray]{0.902}}c}
    \toprule
    Method & Acc.↓ & Comp.↓ & C-$L_1$↓ & NC↑ & F-score↑ \\
    \midrule
    w/o $\mathcal{L}_{nc}$  & \textbf{0.0250} & \textbf{0.0352} & \textbf{0.0301} & 0.9085 & 0.8823 \\
    w/o $\mathcal{L}_{lap}$ & 0.0273 & 0.0362 & 0.0317 & 0.8991 & 0.8728 \\
    w/o $\mathcal{L}_{n}$ & 0.0342 & 0.0362 & 0.0352 & 0.8865 & 0.8199 \\
    w/o Subdivide & 0.0264 & 0.0379 & 0.0321 & \textbf{0.9149} & 0.9052 \\
    w/o Remove & 0.0411	& 0.0363 & 0.0387 & 0.8887 & 0.8197 \\
    Full model & 0.0264 & 0.0399 & 0.0331 & 0.9144 & \textbf{0.9104} \\
    \bottomrule
     \end{tabular}%
    }
    \caption{\textbf{Quantitative results of the ablation study on optimization constraints and mesh refinement strategy.} The best results are marked in \textbf{bold}. The F-score is considered the most convincing metric, as it jointly accounts for both accuracy and completeness, providing a more comprehensive and intuitive evaluation of the overall geometric quality. Our full model consistently achieves the highest F-score, highlighting the effectiveness and complementary benefits of each component in the proposed design.}
    \label{tab:ab_mesh}
\end{table}

\begin{figure}[htb]
  \centering
  \begin{subfigure}[b]{0.49\linewidth}
    \centering
    \includegraphics[width=\linewidth]{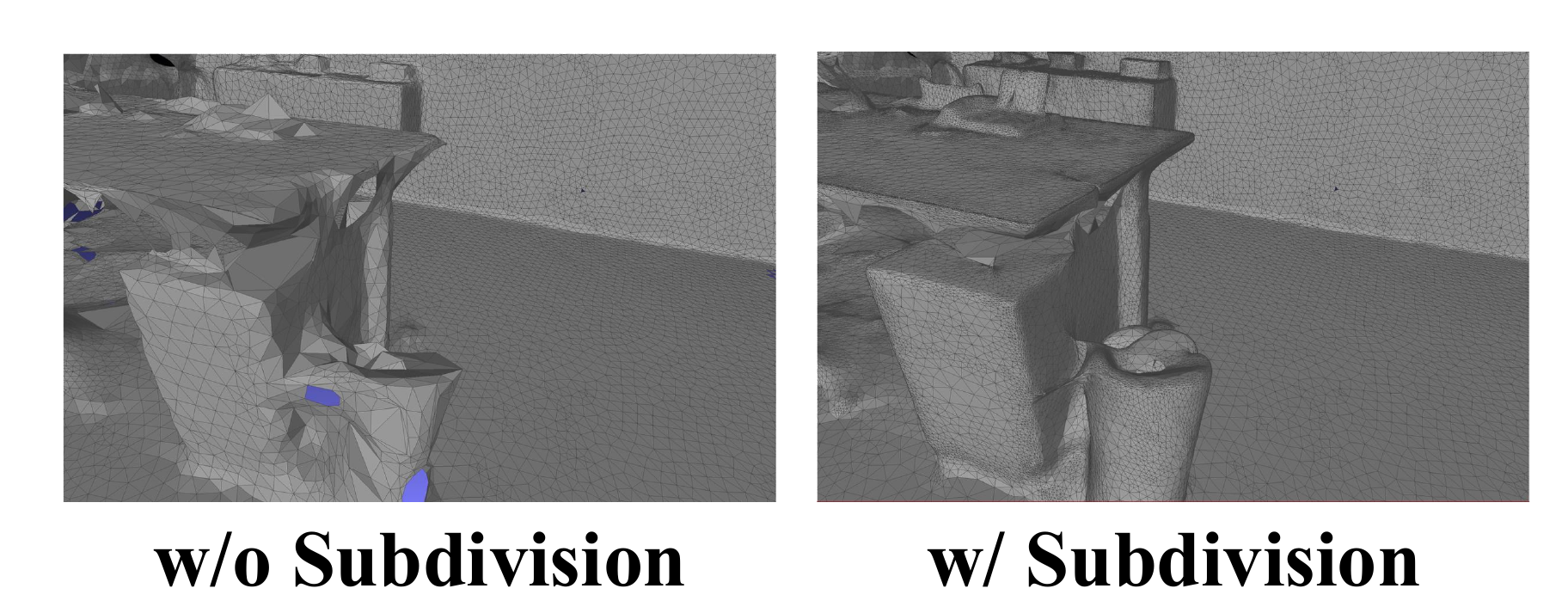}
    \caption{Ablation on subdivision strategy}
    \label{fig:ablation_mesh_refinement_subfig_a}
  \end{subfigure}
  \hfill
  \begin{subfigure}[b]{0.49\linewidth}
    \centering
    \includegraphics[width=\linewidth]{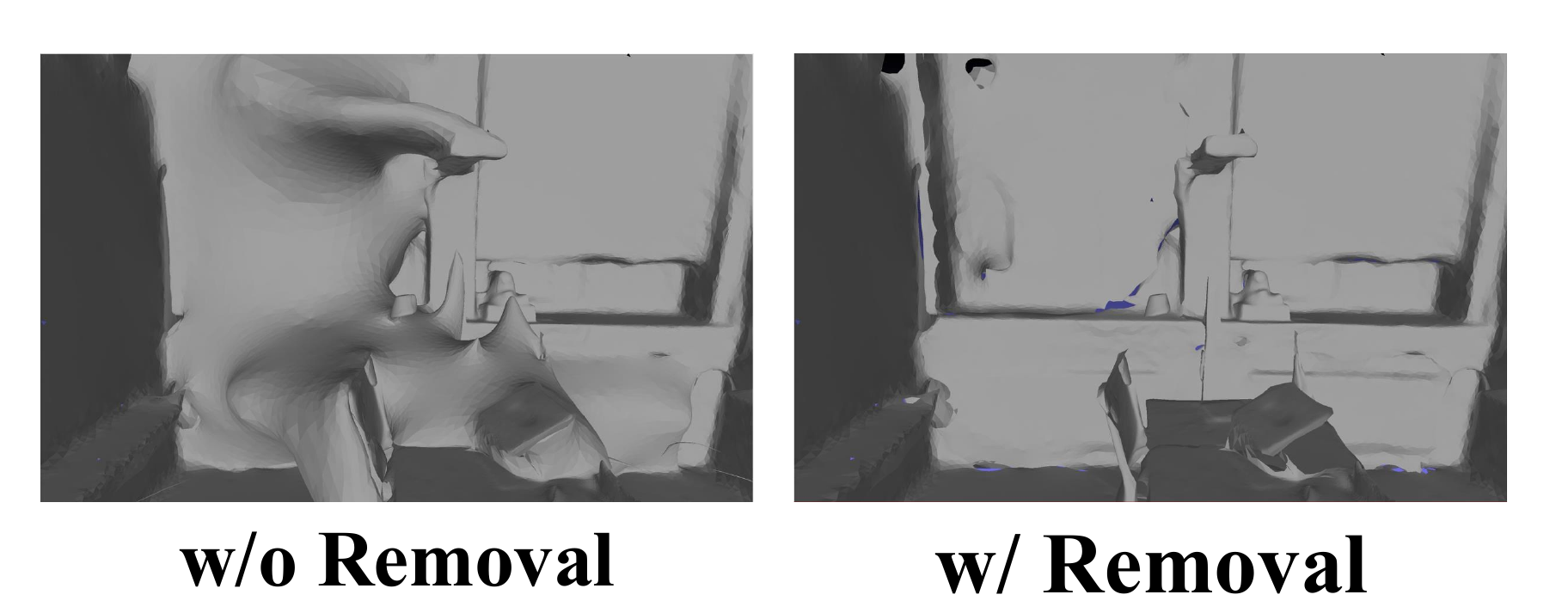}
    \caption{Ablation on removal strategy}
    \label{fig:ablation_mesh_refinement_subfig_b}
  \end{subfigure}
  \caption{\textbf{Qualitative results from ablation studies on mesh refinement strategy.} With the subdivision strategy, the mesh exhibits richer details. Furthermore, the removal strategy effectively eliminates redundant and incorrect mesh faces.}
  \label{fig:abalation_mesh_refinement}
\end{figure}

\noindent\textbf{Mesh Refinement Strategy.} To validate the effectiveness of the proposed mesh refinement design, we conduct an ablation study on the mesh subdivision strategy and the mesh removal strategy, respectively. The quantitative results are shown in \cref{tab:ab_mesh}, and we provide a visualization result in \cref{fig:abalation_mesh_refinement}. Notably, the mesh subdivision strategy significantly improves the intricate details in the scene by splitting mesh faces to fit complex geometry structures. We also provide a further analysis of the effectiveness of the subdivision strategy in \cref{appendix:subdivision}. Additionally, our mesh removal strategy prunes faces that are unsupported by the learned splat distribution, eliminating incorrect mesh geometry introduced by the coarse initialization. This improves mesh–splat consistency and restores the true scene topology. These results collectively demonstrate the importance of both strategies in producing high-quality mesh reconstructions. 

\section{Conclusion}
In this paper, we present a joint representation framework that integrates meshes into Gaussian Splatting for indoor scene reconstruction, enabling the splats to conform closely to the surface while retaining sufficient expressive power. Starting from a coarse and even noisy mesh, we initialize 2D splats to mesh faces with a flexible binding strategy. In addition, we propose an iterative mesh refinement strategy to gradually subdivide and remove mesh faces, which further improves the quality and intricate details of the reconstructed mesh. Extensive experiments on indoor scene reconstruction demonstrate the effectiveness of the proposed approach in improving reconstruction quality and solving challenging reconstruction problems such as texture-less reconstruction.

{
    \small
    \bibliographystyle{ieeenat_fullname}
    \bibliography{main}
}

\clearpage
\appendix
\renewcommand{\thesection}{\Alph{section}}
\renewcommand{\thesubsection}{\Alph{section}.\arabic{subsection}}

\setcounter{page}{1}
\maketitlesupplementary
\section{Implementation Details}
\label{appendix:implementation}
\subsection{Mesh Extraction Methods}
We outline the mesh extraction methods used for different baselines in \cref{tab:mesh_extraction_methods}.

\begin{table}[H]
    \centering
    \scalebox{0.9}
    {
    \begin{tabular}
    {lcc}
    \toprule
    Method & Mesh Extraction & Post hoc Extraction\\
    \midrule
    2DGS & TSDF & $\checkmark$ \\
    3DGS & TSDF & $\checkmark$ \\
    GSRec & Poisson & $\checkmark$ \\
    SuGaR & Poisson & $\checkmark$ \\
    GOF & Marching Tetrahedra & $\checkmark$ \\
    PGSR & TSDF & $\checkmark$ \\
    GaMeS & Poisson & $\times$  \\
    OMeGa & Poisson & $\times$ \\
    \bottomrule
    \end{tabular}
    }
    \caption{Mesh extraction methods used in the experiments}
    \label{tab:mesh_extraction_methods}
\end{table}

\subsection{Hyperparameters}
\noindent\textbf{Parameters.} Regarding the coefficient vector of scale, $s_{rel}$, we set its range to [0.5, 1.5], which guarantees that the splats stay on their corresponding surfaces while maintaining flexible representational capacity.

\noindent\textbf{Weights of losses.} We follow the common practice in prior works and set rendering loss coefficient $\lambda_1=0.8$. The coefficients for Laplacian smoothing loss and normal consistency loss are inherently dependent on the scale of the scene. In indoor scenes reconstructed by COLMAP, we empirically observe that setting $\lambda_{lap}=40$ and $\lambda_{nc}=1$, respectively, achieves a good balance between stability and efficiency during training. We further argue that in other scenarios, these hyperparameters can be straightforwardly rescaled according to the scene scale and the granularity of geometric details. Finally, for the monocular normal loss coefficient $\lambda_{n}$, we adopt an empirical setting of 0.5.

\noindent\textbf{Mesh Refinement.} During mesh refinement, we postpone subdivision and removal until after 1000 iterations to ensure stable training in the early stage. Before 15,000 iterations, subdivision and removal are performed every 500 iterations: the top 2\% of triangles ranked by reconstruction error are subdivided, while the removal threshold $\tau_{remove}$ is fixed at 0.01. This strategy is primarily determined by the granularity of the initial mesh and the desired final resolution. In indoor scenes, we find that such default settings are sufficient to achieve satisfactory performance. 

\section{Additional Evaluation Metrics}
\label{appendix:metrics}
We report training time, peak GPU memory, splat count, and the final model's disk storage in \cref{tab:additional_metrics}. While OMeGa trains longer than most baselines—expected given its joint optimization of an explicit mesh, it uses less GPU memory and yields a smaller on-disk model than most MASt3R-initialized baselines.

\begin{table}[H]
    \centering
    \resizebox{\columnwidth}{!}{%
    \begin{tabular}{lcccc}
    \toprule
    Method & Training time [min] & Memory [GB] & \# Splats [$\times 10^5$] & Disk storage [MB] \\
    \midrule
    3DGS &9 & 1.60 & 7.86 & 185.45 
 \\
    2DGS &16&1.86&8.44&212.60  \\
    2DGS$^\dagger$ &33&9.85&17.85&421.33 \\
    2DGS$^\dagger$ + N &50 &9.86 & 18.36&433.34 \\
    2DGS$^\dagger$ + N + E  &71&13.42&18.05&332.17 \\
    GSRec &105&2.61&2.11&64.05  \\
    SuGaR  &182&15.39&11.00&272.87     \\
    GOF$^\dagger$ &65&19.38&14.08&354.76    
\\
    PGSR$^\dagger$ & 51&15.55&9.77&50.76 
     
 \\
    GaMeS$^\dagger$    &  23& 9.36&12.86&318.90 
 \\
    OMeGa$^\dagger$    & 109 & 5.94 & 13.93 & 287.75 
\\
    \bottomrule
    \end{tabular}%
    }
    \caption{Additional metrics for the main experiments on ScanNet. All values are averaged across 8 scenes. $^\dagger$ indicates initialized with MASt3R results.}
    \label{tab:additional_metrics}
\end{table}

\section{More Discussion on illumination and geometry reconstruction}
\label{appendix:illumination}
\begin{figure}[htb]
  \centering
  \begin{subfigure}[b]{0.49 \linewidth}
    \centering
    \includegraphics[width=\linewidth]{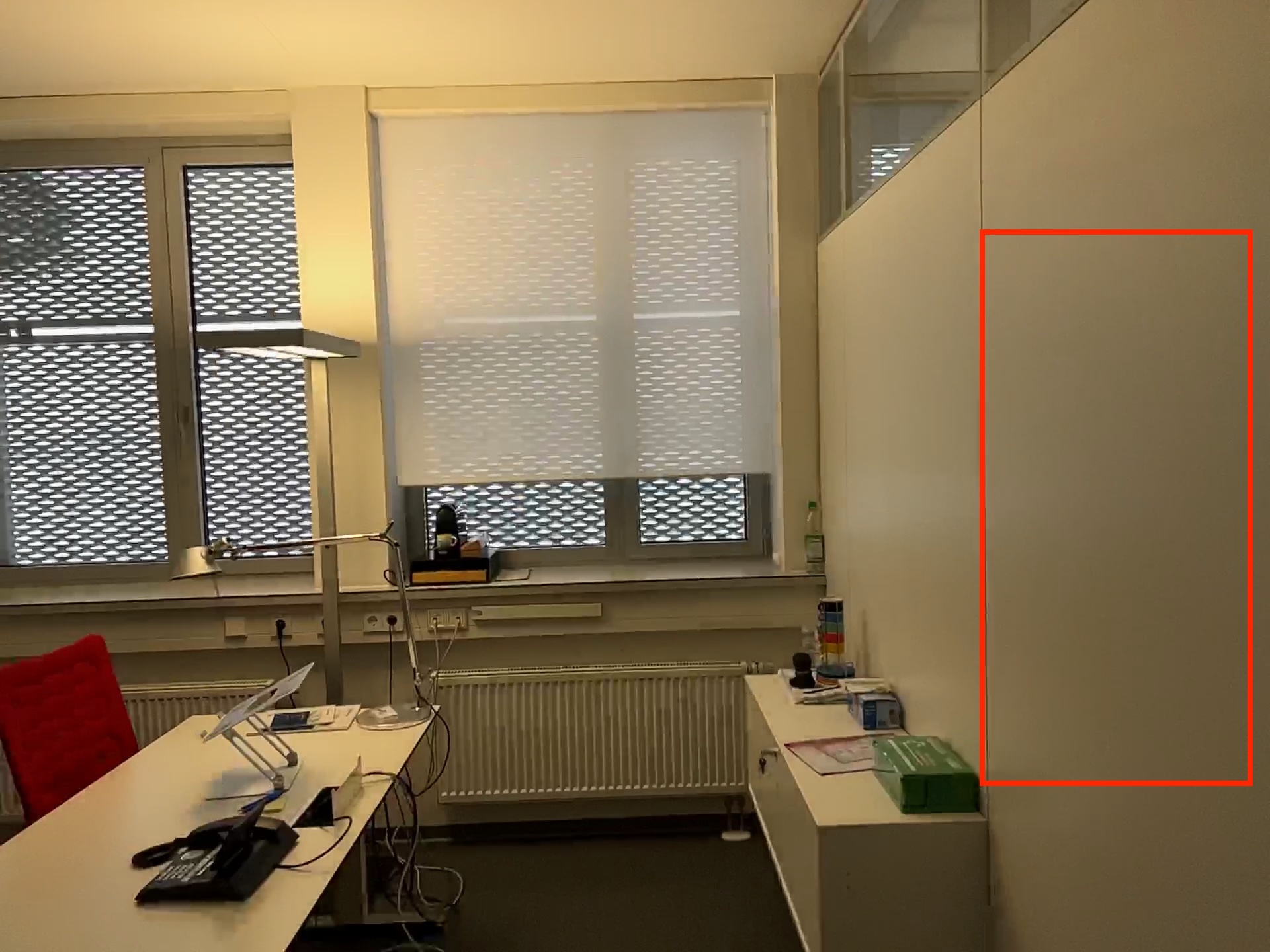}
    \caption{View\#1}
    \label{fig:illumination_example1}
  \end{subfigure}
  \hfill
  \begin{subfigure}[b]{0.49 \linewidth}
    \centering
    \includegraphics[width=\linewidth]{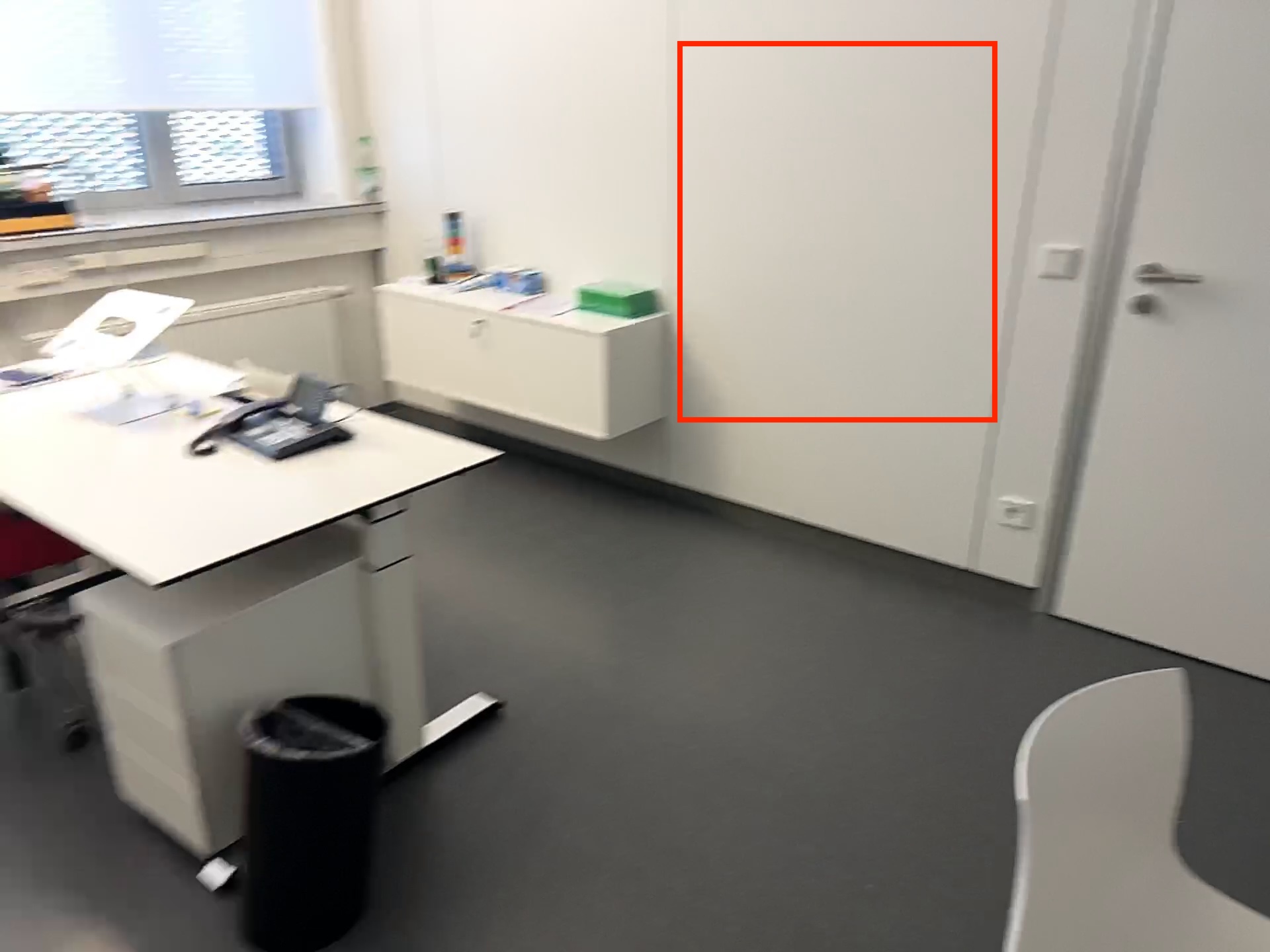}
    \caption{View\#2}
    \label{fig:illumination_example2}
  \end{subfigure}
  \caption{An example of intense illumination change in the ScanNet++ dataset.}
  \label{fig:illum_case}
\end{figure}

\Cref{fig:illum_case} shows a representative indoor scene from ScanNet++ scene \#8b5caf3398 where illumination differs noticeably between two frames. Such variations are common in real-world indoor captures and are known to challenge Gaussian Splatting-based algorithms.

\begin{figure}[htb]
  \centering
  \begin{subfigure}[b]{0.48\linewidth}
    \centering
    \includegraphics[width=\linewidth]{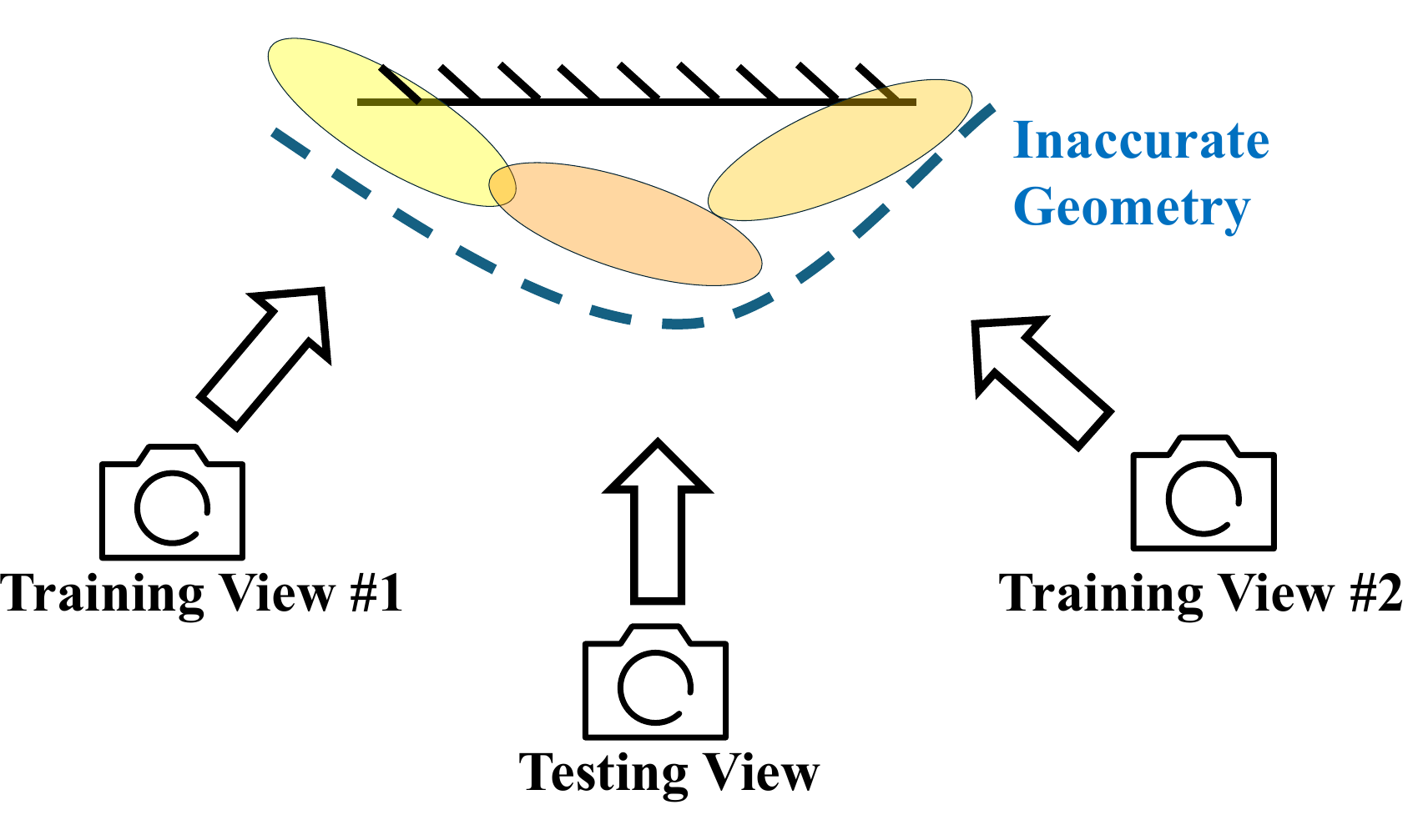}
    \caption{Previous works}
    \label{illumination_discussion_others}
  \end{subfigure}
  \hfill
  \begin{subfigure}[b]{0.48\linewidth}
    \centering
    \includegraphics[width=\linewidth]{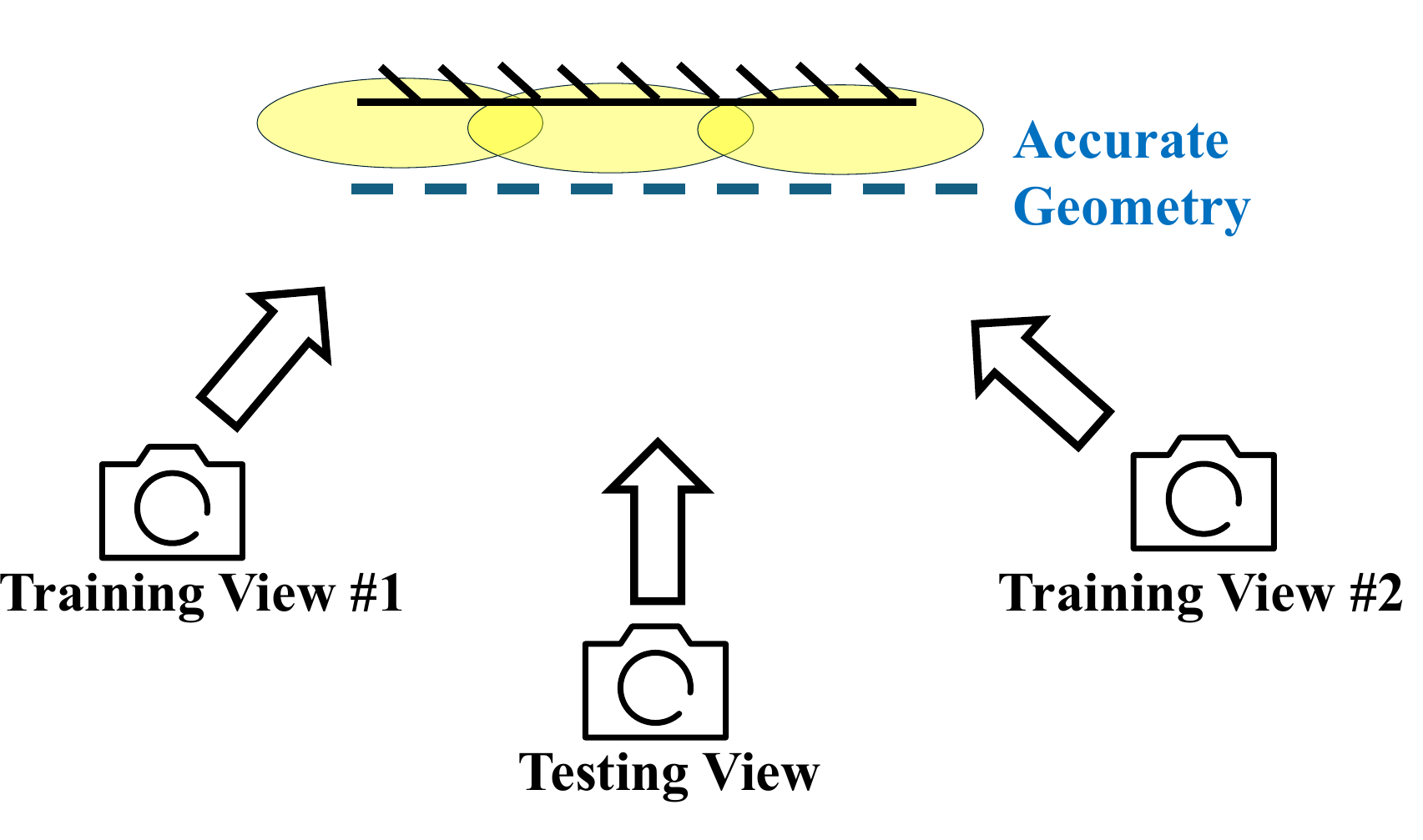}
    \caption{OMeGa}
    \label{fig:illumination_discussion_ours}
  \end{subfigure}
  \caption{Illustration of the difference between OMeGa and previous Gaussian Splatting algorithms in handling illumination changes.}
  \label{fig:illumination_changes_difference}
\end{figure}

As shown in \Cref{fig:illumination_changes_difference}, previous Gaussian Splatting algorithms often overfit the illumination of per-view images by increasing the number of Gaussian Splats in low-texture regions, which leads to multiple layers of Gaussian primitives while the ground truth is a surface (left). This fitting can cause the learned surface to deviate from the true scene, producing artifacts under novel views and inaccurate geometry when extracting the mesh. We identify the previous work's process as \emph{shaping geometry to overfit appearance}.
By contrast, OMeGa anchors splat means and orientations on mesh faces, so \emph{appearance is learned on the correct geometry} rather than reshaping geometry to overfit appearance. Based on the correct geometry, OMeGa captures light variation through feature embeddings (right). As a result, OMeGa achieves both robust novel-view renderings when lighting differs from training views and accurate geometry reconstruction.

\section{Effectiveness of error-based mesh subdivision strategy}
\label{appendix:subdivision}
\begin{figure}[htb]
  \includegraphics[width=\linewidth]{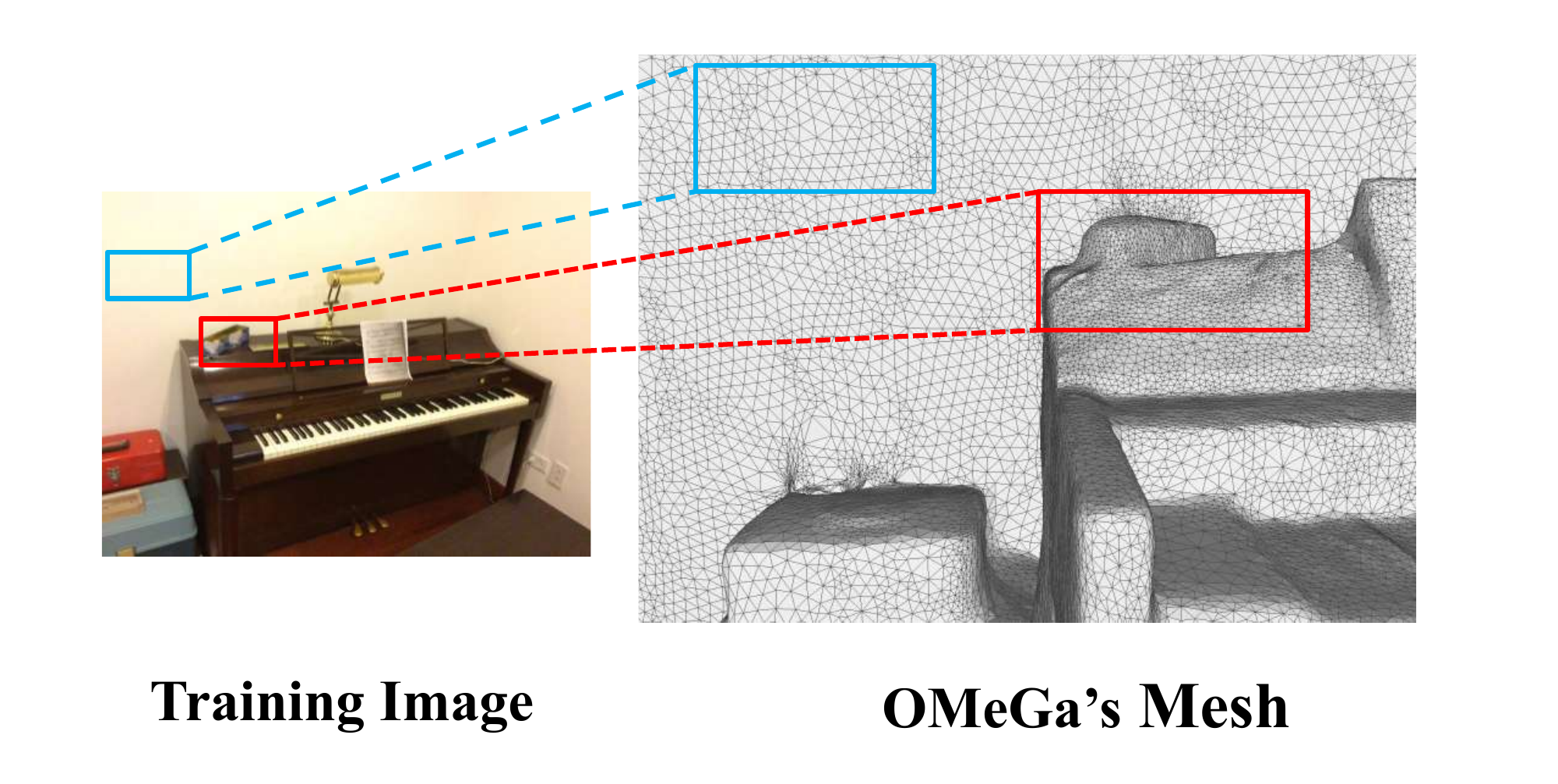}
  \caption{Illustration of the effectiveness of our proposed error-based mesh subdivision strategy.}
  \label{fig:subdivision_example}
\end{figure}

\Cref{fig:subdivision_example} qualitatively demonstrates the effectiveness of our error-based subdivision strategy in adapting to both low-frequency and high-frequency geometry. In textureless, low-frequency regions, the triangle density remains sparse, preserving computational efficiency  (highlighted in the blue box). In contrast, around high-frequency regions (highlighted in the red box), our method iteratively subdivides mesh faces to better fit geometric complexity, leading to a denser and more detailed mesh representation.

\end{document}